\documentclass[lettersize,journal]{IEEEtran}
\usepackage{amsmath,amsfonts}
\usepackage{algorithmic}
\usepackage{algorithm}
\usepackage{array}
\usepackage[caption=false,font=normalsize,labelfont=sf,textfont=sf]{subfig}
\usepackage{textcomp}
\usepackage{stfloats}
\usepackage{url}
\usepackage{verbatim}
\usepackage{graphicx}
\usepackage{cite}

\usepackage{color}
\usepackage{makecell}
\usepackage{multirow}
\usepackage{hyperref}
\usepackage{bm}
\usepackage{bbding}
\usepackage{pifont}
\usepackage{dsfont}
\usepackage{booktabs}
\usepackage{stfloats}
\begin{document}

\title{CorrNet+: Sign Language Recognition and Translation  via Spatial-Temporal Correlation}

\author{Lianyu Hu, Wei Feng$^{\dagger}$, \IEEEmembership{Member,~IEEE}, Liqing Gao, Zekang Liu, Liang Wan$^{\dagger}$, \IEEEmembership{Member,~IEEE}
\thanks{Lianyu Hu, Wei Feng, Liqing Gao, Zekang Liu, Liang Wan are with the College of Intelligence and Computing, Tianjin University, Tianjin
300350, China (e-mail: hly2021@tju.edu.cn; wfeng@ieee.org; lwan@tju.edu.cn).

$\dagger$ Wei Feng and Liang Wan are the Corresponding authors.}}

\markboth{Journal of \LaTeX\ Class Files,~Vol.~14, No.~8, August~2021}%
{Shell \MakeLowercase{\textit{Hu et al.}}: CorrNet+: Sign Language Recognition and Translation via Spatial-Temporal Correlation}


\maketitle

\begin{abstract}
In sign language, the conveyance of human body trajectories predominantly relies upon the coordinated movements of hands and facial expressions across successive frames. Despite the recent impressive advancements of sign language understanding methods, they often solely focus on individual frames, inevitably overlooking the inter-frame correlations that are essential for effectively modeling human body trajectories. To address this limitation, this paper introduces a spatial-temporal correlation network, denoted as CorrNet+, which explicitly identifies and captures body trajectories across multiple frames. In specific, CorrNet+ employs two parallel modules to build human body trajectories: a correlation module and an identification module. The former captures the cross-spacetime correlations in local spatial-temporal neighborhoods, while the latter dynamically constructs human body trajectories by distinguishing informative spatial regions. Afterwards, a temporal attention module is followed to adaptively evaluate the contributions of different frames in the whole video. The resultant features offer a holistic perspective on human body movements, facilitating a deeper understanding of sign language. As a unified model, CorrNet+ achieves new state-of-the-art performance on two extensive sign language understanding tasks, including continuous sign language recognition (CSLR) and sign language translation (SLT). Especially, CorrNet+ surpasses previous methods equipped with resource-intensive pose-estimation networks or pre-extracted heatmaps for hand and facial feature extraction. Compared with CorrNet, CorrNet+ achieves a significant performance boost across all benchmarks while halving the computational overhead, achieving a better computation-accuracy trade-off. A comprehensive comparison with previous spatial-temporal reasoning methods verifies the superiority of CorrNet+. Code is available at \url{https://github.com/hulianyuyy/CorrNet_Plus}. 

\end{abstract}


\section{Introduction}

\begin{figure*}[t]
  \centering
  \includegraphics[width=\linewidth]{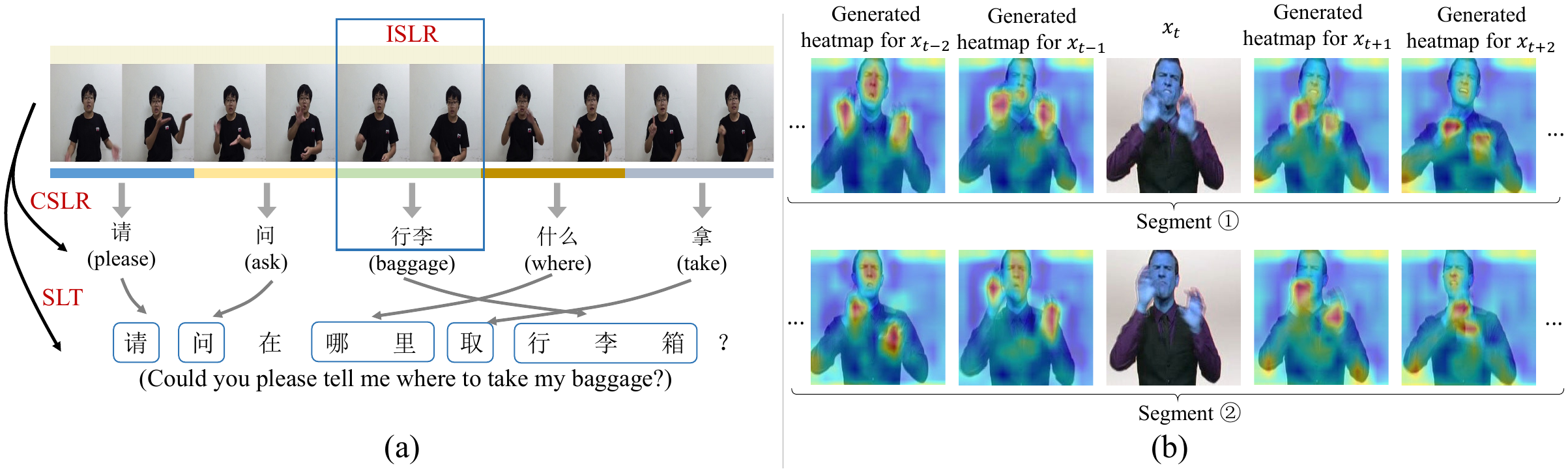}
  \caption{(a) Illustration for the difference among the isolated sign language recognition (ISLR) task, continuous sign language recognition (CSLR) task and sign language translation (SLT) task. (b) Visualization of correlation maps with Grad-CAM~\cite{selvaraju2017grad} between the current frame and two adjacent frames in the left/right side. It's observed that without extra supervision, our method well attends to informative regions in adjacent frames to identify human body trajectories. }
  \label{fig1}
\end{figure*}

Sign language is one of the most widely-used communication tools for the deaf community in their daily life, which mainly conveys its meaning by facial expressions, head movements, hand gestures and body postures~\cite{dreuw2007speech,ong2005automatic}. However, mastering this language remains an overwhelming challenge for the hearing people, thus hindering direct interactions between two distinct groups. To alleviate this barrier, recent strides in automatic sign language understanding techniques~\cite{adaloglou2021comprehensive,rastgoo2021sign} have emerged, broadly categorized into three distinct domains: (1) isolated sign language recognition (ISLR), which aims to classify a video segment into an independent gloss\footnote{Gloss is the atomic lexical unit to annotate sign languages.}; (2) continuous sign language recognition (CSLR), which progresses by classifying the input sign videos into a series of glosses to express sentences, instead of recognizing a single gloss only; (3) sign language translation (SLT), which directly translating the input sign videos into spoken texts that can be naturally understood by the hearing people. The difference of these tasks is illustrated in Fig.~\ref{fig1}(a). To hopefully bridge the communication gaps between two groups, this paper focuses on CSLR and SLT, as they hold greater promise for real-life applications in sign language systems.

\vspace{-2px}

Evidently, human body trajectories serve as prominent cues for understanding actions in human-centric video comprehension, which have gained substantial attention across various  tasks~\cite{carreira2017quo,wang2018temporal,shou2016temporal,weinzaepfel2015learning,zhu2017uncovering,li2022invariant}. In sign language, these trajectories are mainly conveyed by both manual components (hand/arm gestures), and non-manual components (facial expressions, head movements, and body postures)~\cite{dreuw2007speech,ong2005automatic}. Especially, the coordinated horizontal and vertical movements of human face and both hands, coupled with adjoint actions like finger twisting and facial expressions, play a major role in expressing sign language. Tracking and leveraging the trajectories of these crucial body parts is of great benefit to understanding sign language. 

However, current sign language methods~\cite{cheng2020fully,cui2019deep,niu2020stochastic,Min_2021_ICCV,zuo2022c2slr,hao2021self,hu2023self,zhou2023gloss,guo2023distilling,zheng2023cvt} usually treat each frame equally, overlooking their  cross-frame interactions and thereby failing to leverage human body trajectories. Especially, they usually adopt a shared 2D CNN to independently extract spatial features for each frame~\cite{zheng2023cvt,cheng2020fully,Min_2021_ICCV,guo2023distilling,hu2023self}. Consequently, frames are processed individually without considering their interactions, thus inhibited to harness the potential of cross-frame trajectories for sign comprehension. Some methods propose to use a 3D or (2+1)D CNN~\cite{cui2019deep,chen2022two} to capture the local cross-spacetime features. However, their fixed design and limited spatial-temporal receptive fields hinder the establishment of spatial relationships across distant regions. Moreover, these methods incur substantial computational costs compared to their 2D counterparts. Alternative temporal techniques, such as temporal shift~\cite{lin2019tsm} or temporal convolutions~\cite{liu2020teinet}, could address short-term temporal dynamics. However, it's hard for them to aggregate information from distant spatial regions due to the limited spatial-temporal receptive field. Besides, they may fail to dynamically model human body movements for different samples with a fixed structure during inference. With the above considerations, it's necessary to develop an effective and efficient method for capturing human body trajectories to advance sign language comprehension. 

To address these challenges, we introduce CorrNet+, a novel framework explicitly designed to model human body trajectories across adjacent frames. As depicted in Figure 1(b), our approach dynamically attends to the movements of informative regions across wide spatial distances. Unlike certain prior methods~\cite{zuo2022c2slr,zhou2020spatial,chen2022two,koller2019weakly} that rely on expensive supervision such as pose estimation techniques or body heatmaps, our method alleviates the need for such resource-intensive guidance and can be trained in a self-motivated manner. Notably, our approach achieves superior performance compared to previous methods while significantly reducing the required computational demands.

CorrNet+ employs two parallel modules to build human body trajectories: a correlation module and an identification module. The former computes correlation maps within a local spatial-temporal region to identify human body trajectories. The latter dynamically emphasizes the informative regions that convey critical information. Besides these two components, considering human body trajectories are unevenly distributed in the video, a temporal attention module is then introduced to highlight the critical human body movements. The generated features provide a comprehensive perspective on human body movements, thereby enhancing the comprehension of sign language. Remarkably, CorrNet+ achieves new state-of-the-art performance on three large-scale CSLR benchmarks (PHOENIX2014~\cite{koller2015continuous}, PHOENIX2014-T~\cite{camgoz2018neural} and CSL-Daily~\cite{zhou2021improving}), and two widely-used SLT benchmarks (PHOENIX2014-T~\cite{camgoz2018neural} and CSL-Daily~\cite{zhou2021improving}). Especially, CorrNet+ largely outperforms previous methods equipped with resource-intensive pose-estimation networks or pre-extracted heatmaps for hand and facial feature extraction~\cite{zuo2022c2slr,zhou2020spatial,chen2022two,koller2019weakly}. Compared with CorrNet~\cite{hu2023continuous}, CorrNet+ brings notable performance gain across all benchmarks and drastically reduces the consumed computations by half, achieving a better computation-accuracy trade-off. A comprehensive comparison with other spatial-temporal reasoning methods demonstrates the superiority of CorrNet+. Visualizations hopefully verify the efficacy of CorrNet+ on emphasizing human body trajectories across adjacent frames. Abundant ablations demonstrate the effects of each component within CorrNet+. 

This paper is a substantial extension from a preliminary conference version~\cite{hu2023continuous} with a number of major changes. First, we reformulate the design of the correlation module in Section 3.2 to make it more lightweight and powerful, which is a key component for effectively modeling human body trajectories. Second, a new temporal attention module is introduced to dynamically emphasize the critical body trajectories in Section 3.4. Finally, we incorporate new results on the SLT benchmarks, and significantly extend the experimental results on the CSLR benchmarks in Section 4. We additionally append new visualizations to clearly show the effects of our proposed method. The remainder of this paper is organized as follows. Section 2 reviews the related work. Section 3 elaborates the proposed method. Section 4 reports the experimental results, followed by a brief conclusion in Section 5. 

\section{Related Work}
\subsection{Continuous Sign Language Recognition}
Continuous sign language recognition tries to translate image frames into corresponding glosses in a weakly-supervised way: only sentence-level label is provided. Earlier methods~\cite{gao2004chinese,freeman1995orientation} in CSLR always employ hand-crafted features or HMM-based systems~\cite{koller2016deepsign,han2009modelling,koller2017re,koller2015continuous} to perform temporal modeling and translate sentences step by step. Hand-crafted features~\cite{gao2004chinese,freeman1995orientation} are carefully selected to provide better visual information. HMM-based systems\cite{koller2016deepsign,han2009modelling,koller2017re,koller2015continuous} first employ a feature extractor to capture visual features and then adopt an HMM to perform long-term temporal modeling. 

The recent success of convolutional neural networks (CNNs) and recurrent neural networks (RNNs) brings huge progress for CSLR.  The widely used CTC loss~\cite{graves2006connectionist} in recent CSLR methods~\cite{pu2019iterative,pu2020boosting,cheng2020fully,cui2019deep,niu2020stochastic,Min_2021_ICCV} enables training deep networks in an end-to-end manner by sequentially aligning target sentences with input frames. These CTC-based methods first rely on a feature extractor, i.e., 3D or 2D\&1D CNN hybrids, to extract frame-wise features, and then adopt a LSTM for capturing long-term temporal dependencies. However, several methods~\cite{pu2019iterative,cui2019deep} found in such conditions the feature extractor is not well-trained and then present an iterative training strategy to relieve this problem, but consume much more computations. Some recent studies~\cite{Min_2021_ICCV,cheng2020fully,hao2021self,min2022deep} try to directly enhance the feature extractor by adding alignment losses~\cite{Min_2021_ICCV,hao2021self,min2022deep} or adopt pseudo labels~\cite{cheng2020fully} in a lightweight way, alleviating the heavy computational burden. TLP~\cite{hu2022temporal} proposes to enhance the temporal information extraction process by designing advanced temporal pooling methods. SEN~\cite{hu2023self} tries to locate the informative spatial regions in sign videos in a self-supervised way. CVT-SLR~\cite{zheng2023cvt} employs a contrastive visual-textual transformation to tackle the insufficient training problem existed in CSLR. CTCA~\cite{guo2023distilling} designs a cross-temporal context aggregation module to enhance local temporal context and global temporal context.

Our method is designed to explicitly incorporate body trajectories to identify a sign, especially those from hands and face. Some previous methods have also explicitly leveraged the hand and face features for better recognition. For example, CNN-LSTM-HMM~\cite{koller2019weakly} employs a multi-stream HMM (including hands and face) to integrate multiple visual inputs to improve recognition accuracy. STMC~\cite{zhou2020spatial} first utilizes a pose-estimation network to estimate human body keypoints and then sends cropped appearance regions (including hands and face) for information integration. C$^2$SLR~\cite{zuo2022c2slr} leverages the pre-extracted pose keypoints as supervision to guide the model to explicitly focus on hand and face regions. TwoStream Network~\cite{chen2022two} builds two branches consisting of a visual branch and a pose branch to fuse beneficial information from complementary modalities. Our method doesn't rely on additional cues like heavy pose estimation networks~\cite{zuo2022c2slr,zhou2020spatial,chen2022two} or multiple streams~\cite{koller2019weakly} which consume much more computations to leverage hand and face information. Instead, our model could be end-to-end trained to dynamically attend to body trajectories in a self-motivated and lightweight way.

\subsection{Sign Language Translation}
Camgoz et al.~\cite{camgoz2018neural} pioneer the neural SLT task and publish the neural dataset PHOENIX2014-T~\cite{camgoz2018neural} which regards the SLT as a sequence-to-sequence problem. They implement the neural SLT system using the encoder-decoder paradigm~\cite{bahdanau2014neural}. This paradigm is adopted by subsequent studies which focus on addressing the challenges of data scarcity and domain gap. Then, SLRT~\cite{camgoz2020sign} first introduces a Transformer-based encoder-decoder framework to perform end-to-end SLT, with a Connectionist Temporal Classification (CTC) loss~\cite{graves2006connectionist} to soft-match sign representations and gloss sequences. STMC-T~\cite{zhou2020spatial} improves sign language translation by introducing multiple cues aimed by a pose estimation network. SignBack~\cite{zhou2021improving} tries to handle the insufficient training data problem by introducing back-translation techniques to generate new pseudo samples. Motivated by the progress of neural machine translation (NMT), several methods attempt to introduce these advanced techniques into SLT. For example, Chen et al.~\cite{chen2022simple,chen2022two} made the first attempt to introduce large language models into SLT with carefully designed pretraining strategies. XmDA~\cite{ye2023cross} presents two new data augmentation methods, namely, cross-modality mix-up and cross-modality knowledge distillation to expand the training samples. Zhu el al.~\cite{zhu2023neural} testifies the effectiveness of several NMT techniques including data augmentation, transfer learning and multilingual NMT on SLT. Most existing methods adopt gloss representations as an intermediate state to promote translation accuracy. Some methods~\cite{lin2023gloss,zhou2023gloss} propose to eliminate the need of label-laboring glosses and design gloss-free SLT methods. GloFE~\cite{lin2023gloss} presents an end-to-end sign language translation framework by exploiting the shared underlying semantics of signs and the corresponding spoken translation. GFSLT-VLP~\cite{zhou2023gloss} improves SLT by inheriting language-oriented prior knowledge from pretrained models, without any gloss annotation assistance.

\subsection{Applications of Correlation Operation}
Correlation operation has been widely used in various domains, especially optical flow estimation and video action recognition. Rocco et al.~\cite{rocco2017convolutional} used it to estimate the geometric transformation between two images, and Feichtenhofer et al.~\cite{feichtenhofer2017detect} applied it to capture object co-occurrences across time in tracking. For optical flow estimation, Deep matching~\cite{weinzaepfel2013deepflow} computes the correlation maps between image patches to find their dense correspondences. CNN-based methods like FlowNet~\cite{dosovitskiy2015flownet} and PWC-Net~\cite{sun2018pwc} design a correlation layer to help perform multiplicative patch comparisons between two feature maps. More recently, VideoFlow~\cite{shi2023videoflow} proposes to propagate motion correlations between adjacent frames for multi-frame optical flow estimation. FlowFormer++~\cite{shi2023flowformer++} introduces a masked autoencoding pretraining strategy and encodes the cross-frame correlations to help optical flow estimation. For video action recognition, Zhao et al.~\cite{zhao2018recognize} firstly employ a correlation layer to compute a cost volume to estimate the motion information. STCNet~\cite{diba2018spatio} considers spatial correlations and temporal correlations, respectively, inspired by SENet~\cite{hu2018squeeze}. MFNet~\cite{lee2018motion} explicitly estimates the approximation of optical flow based on fixed motion filters. Wang et al.~\cite{wang2020video} design a learnable correlation filter and replace 3D convolutions with the proposed filter to capture spatial-temporal information. PCD~\cite{xu2022aligning} presents to minimize the distribution of correlation information in videos for domain adaptation. Different from these methods that explicitly or implicitly estimate optical flow, the correlation operator in our method is used in combination with other operations to identify and track body trajectories across frames. 

\section{Method}
\subsection{Overview}
\begin{figure*}[t]
  \centering
  \includegraphics[width=0.85\linewidth]{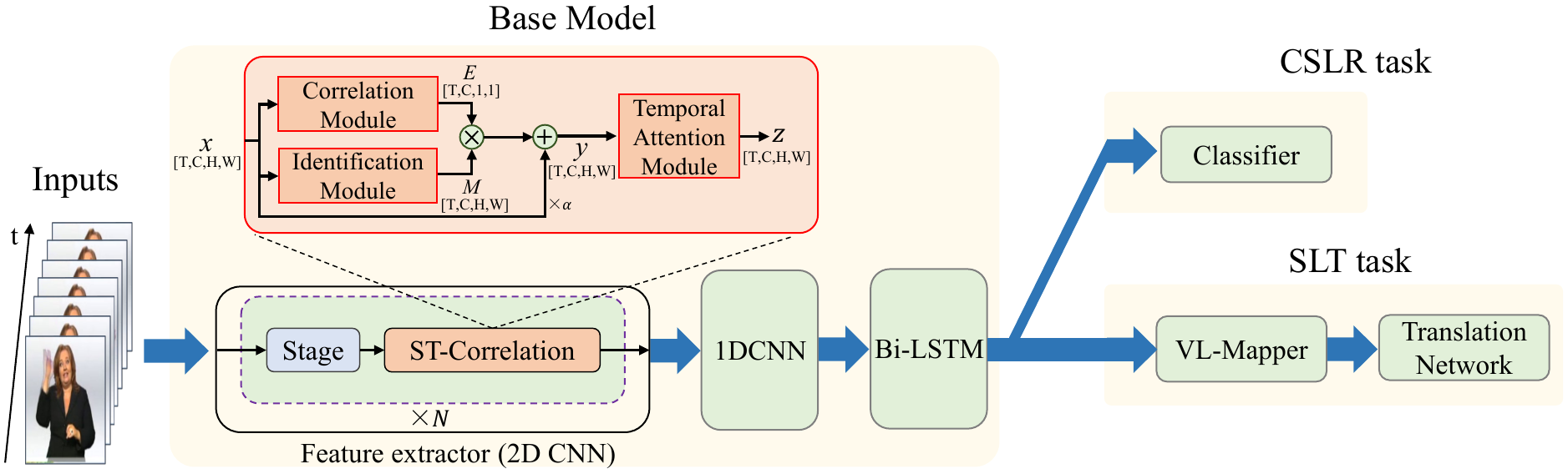}
  \caption{An overview for our CorrNet+, which can support both the CSLR task and the SLT task with a common base model. In this base model, it first employs a feature extractor (2D CNN) to capture frame-wise features, and then adopts a 1D CNN and a BiLSTM to perform short-term and long-term temporal modeling, respectively. For the CSLR task, we attach a classifier instantiated as a fully connected layer to perform classification. For the SLT task, we attach a VL-mapper instantiated as a MLP and a translation network to predict sentences. The feature extractor is consisted of multiple stages to extract spatial-wise features for each frame independently. After each stage of the feature extractor, we insert a correlation stage to capture cross-frame interactions. An identification module and a correlation module are first concurrently placed to identify body trajectories across adjacent frames, whose outputs are then element-wisely multiplied and fed into the temporal attention module to dynamically emphasize the key human body trajectories in the whole video.}
  \label{fig2}
  \end{figure*}

As shown in Fig.~\ref{fig2}, our model comprises a foundational base model, followed by different task-specific heads to support various sign language understanding tasks. Given a sign video with $T$ input frames $\bm{x} = \{\bm{x}^{0}_{t}\}_{t=1}^T \in \mathcal{R}^{T \times 3\times H_0 \times W_0} $ with spatial size of $H_0\times W_0$, the base model first uses a feature extractor instantiated as a 2D CNN\footnote{Here we only consider the feature extractor based on 2D CNN, because recent findings~\cite{adaloglou2021comprehensive,zuo2022c2slr} show 3D CNN can not provide as precise gloss boundaries as 2D CNN, and lead to lower accuracy. } to extract spatial-wise features $\bm{v} = \{\bm{v}_t\}_{t=1}^{T} \in \mathcal{R}^{T\times d}$ with $d$ representing the number of channels. It further incorporates a 1D CNN and a BiLSTM to perform short-term and long-term temporal modeling, respectively. Various task-specific heads are attached to support different sign language understanding tasks. For the CSLR task, we attach a classifier instantiated as a fully connected layer to recognize the input video into a series of glosses $\bm{g}=\{ \bm{g}_i\}_{i=1}^{N}$. Here, $N$ denotes the length of the label sequence. This process is supervised by the widely-used CTC loss~\cite{graves2006connectionist} $\mathcal{L}_{\rm CTC}$ to align input video frames with target gloss sequences. For the SLT task, we attach a visual-language (VL) mapper instantiated as a MLP and a translation network to translate the gloss-wise features $v$ into spoken texts $\bm{s}=\{ \bm{s}_i\}_{i=1}^{H}$. Here, $H$ denotes the length of the output text sequence. This procedure is supervised by the standard sequence-to-sequence cross-entropy loss~\cite{vaswani2017attention} $\mathcal{L}_{\rm CE}$.

\vspace{-5px}
Despite the recent advancements in sign language understanding methods, they usually treat each frame equally by using a common 2D CNN to extract spatial-wise features and thus fail to capture cross-frame interactions. While some methods propose to model local spatial-temporal information with spatial-temporal reasoning methods like 3D CNN~\cite{cui2019deep,chen2022two} and temporal convolutions, they suffer from excessive computations and limited spatial-temporal receptive fields. Consequently, they struggle to effectively capture human body movements across a broader spatial-temporal region. To address these limitations, we design a spatial-temporal correlation network (CorrNet+) as shown in Fig.~\ref{fig2}. We seamlessly insert a spatial-temporal correlation network (ST-correlation) after each stage in the feature extractor to capture the local spatial-temporal correlations for each frame. Specifically, We simultaneously deploy two critical components including a correlation module and an identification module to capture the cross-frame interactions and identify informative spatial regions. The outputs $\bm{E}$ and $\bm{M}$ from both modules are element-wisely multiplied and then added via a residual input connection, yielding intermediate representations $\bm{y}$. We then feed $\bm{y}$ into a temporal attention module to dynamically evaluate the contributions of different frames in the whole video to emphasize keyframes and suppress meaningless ones. We next introduce each component in detail.
\subsection{Correlation Module}
As a rich and expressive communication protocol, sign language is mainly conveyed by both manual components (hand/arm gestures), and non-manual components (facial expressions, head movements, and body postures)~\cite{dreuw2007speech,ong2005automatic}. However, these informative body parts, e.g., hands and face, often exhibit misalignment across adjacent frames. To address this spatial discrepancy and establish connections between distant spatial regions, we propose a novel approach by computing correlation maps between neighboring frames to identify and track human body trajectories. We first briefly recap the solution of CorrNet~\cite{hu2023continuous} and naturally introduce our solution to overcome its inherent limitations.

\begin{figure*}[t]
  \centering
  \includegraphics[width=0.85\linewidth]{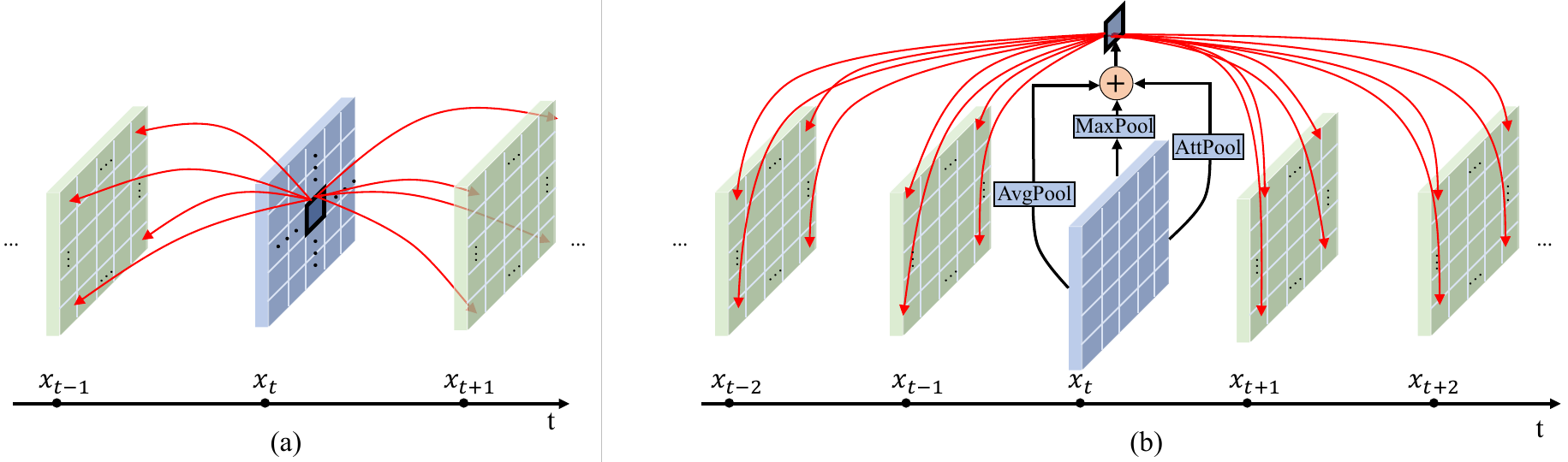}
  \caption{Illustration for the difference between the correlation operator in CorrNet~\cite{hu2023continuous} and CorrNet+. (a) CorrNet~\cite{hu2023continuous}. It computes correlation maps between a spatial patch $p_t(i,j)$ in $x_t$ and all other patches in adjacent frame $x_{t+1}$ and $x_{t-1}$. The overall computation complexity is $O(H^2W^2)$, quadratic to the number of spatial patches in each frame, which incurs heavy extra computations. (b) To reduce computations, we condense the features of $x_t$ into several compact representations, which are then used to compute correlation maps with adjacent frames on behalf of $x_t$. In this case, as the number of selected patches is reduced from $H\times W$ to $O(1)$ for $x_t$, the computation complexity is drastically decreased from $O(H^2W^2)$ to $O(HW)$. It also enables us to compute correlation maps with neighbors in a larger temporal duration to more effectively capture the whole human body movements in expressing a sign.}
  \label{fig3}
  \end{figure*}

Formally, each frame could be represented as a 3D tensor $\bm{x}_t \in \mathcal{R}^{C\times H \times W}$, where $C$ represents the number of channels and $H\times W$ denotes spatial size. In CorrNet~\cite{hu2023continuous}, we compute the affinities between all patches in the current frame $\bm{x}_t$ and patches in adjacent frames to model human body trajectories. Taking a feature patch $\bm{p}_t(i,j)$ with the spatial location $(i,j)$ in the current frame $\bm{x}_t$ as an example, its affinity $\bm{A}(i,j,i',j')$ with another patch $\bm{p}_{t+1}(i',j')$ in $\bm{x}_{t+1}$ is computed in a dot-product way as: 
\begin{equation}
  \label{e1}
  \bm{A}(i,j,i',j') = \frac{1}{C} \sum_{c=1}^{C}{ \bm{p}^c_t(i,j) \times \bm{p}^c_{t+1}(i',j').}
\end{equation}
Fig.~\ref{fig3}(a) illustrates this process. However,  the computed correlation maps yield a tensor of size $H\times W\times H \times W$, resulting in an overall computation complexity of $O(H^2W^2)$ quadratic to the number of patches. Though this operation can effectively build cross-frame interactions to handle the spatial misalignment, it imposes a substantial computational burden. Moreover, the high computational costs restrict the spatial-temporal interactions to neighboring frames, hindering our ability to consecutively capture human body trajectories across a broader temporal context.

\begin{figure}[t]
  \centering
  \includegraphics[width=\linewidth]{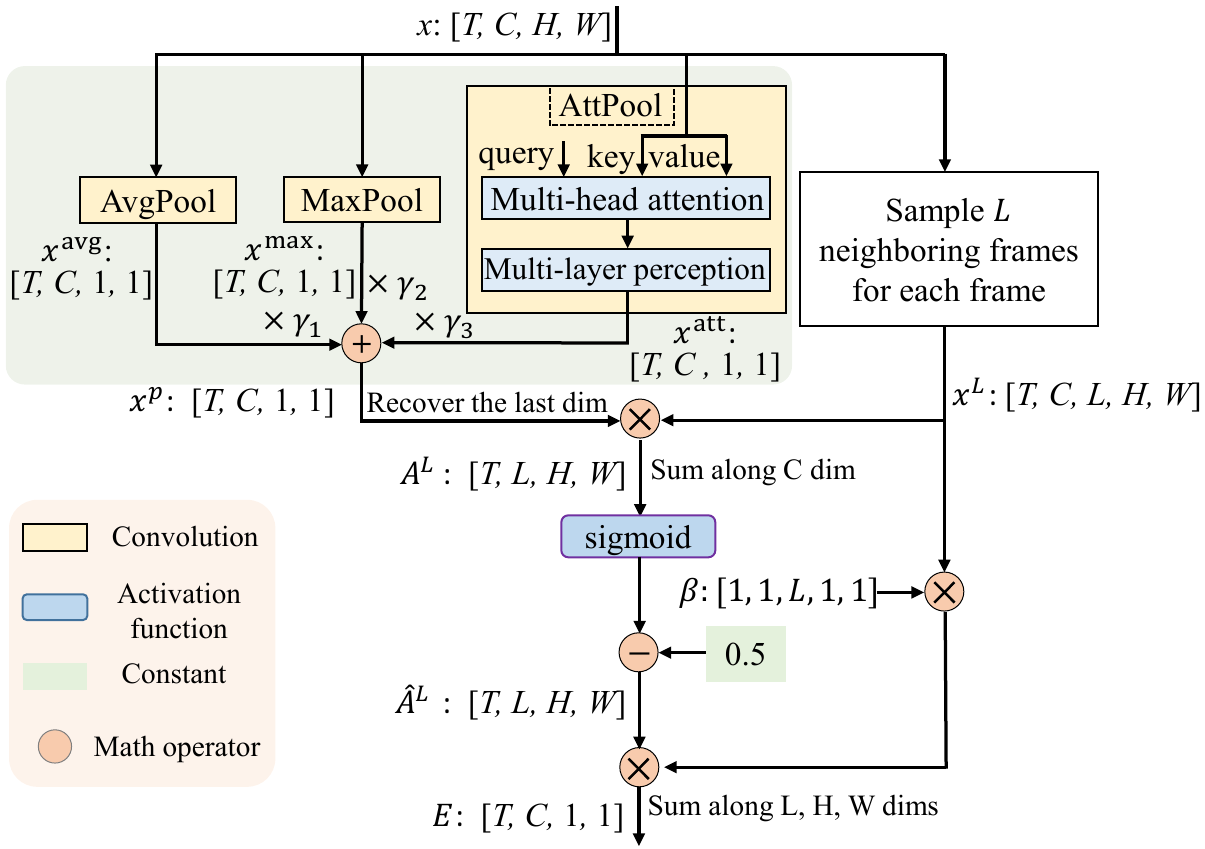}
  \caption{An framework overview for our proposed correlation module. It first condenses each frame into a compact representation, and then uses it to compute correlation maps with adjacent frames within a predefined range of $L$ to model human body trajectories.}
  \label{fig3_main}
  \end{figure}
To handle these limitations, we reformulate the correlation module to make it more lightweight and powerful, whose framework is shown in fig.~\ref{fig3_main}. Specifically, we compress all patches in $\bm{x}_t$ into a compact tensor to compute the correlation maps with significantly reduced computational overhead. We further extend the spatial-temporal neighborhood of the correlation operator to capture the trajectories of the signer in a large temporal duration.

In specific, we use three different ways to compress the features of each frame from various views. For simplicity, we choose the average aggregation, maximum aggregation and attention aggregation functions as our protocols. 

For average aggregation, given the input feature $\bm{x} \in \mathcal{R}^{T\times C\times H \times W}$, we perform average pooling along the spatial dimension to transform it into a representation $\bm{x}^{\rm avg}\in \mathcal{R}^{T\times C\times 1 \times 1}$ as:
\begin{equation}
  \label{e2}
  \bm{x}^{\rm avg} = {\rm AvgPool}(\bm{x}) .
\end{equation}

For maximum aggregation, we perform max pooling to compress $\bm{x}$ into a representation $\bm{x}^{\rm max}\in \mathcal{R}^{T\times C\times 1 \times 1}$ as:
\begin{equation}
  \label{e3}
  \bm{x}^{\rm max} = {\rm MaxPool}(\bm{x}).
\end{equation}

For attention aggregation, we randomly initialize a tensor $\bm{q}\in \mathcal{R}^{1\times C\times 1 \times 1}$ acting as a query. It is then used to compute affinities $\bm{A}\in \mathcal{R}^{T \times 1\times H\times W}$ with patches in each frame following the multi-head attention (MHA)~\cite{vaswani2017attention} process, whose features are fed into a Multi-Layer Perception (MLP) module~\cite{vaswani2017attention} to obtain the output $\bm{x}^{\rm att} \in \mathcal{R}^{T\times C\times 1 \times 1}$ as: 
\begin{equation}
  \label{e4}
  \bm{x}^{\rm att} = {\rm MLP}({\rm MHA}({\rm query}=\bm{q}, {\rm key}=\bm{x}, {\rm value}=\bm{x})) .
\end{equation}

In this procedure, the number of heads is set as 1 for the MHA process, and the dimension expansion factor is 1 for the MLP module to minimize computations.  

After obtaining the condensed features $\bm{x}^{\rm avg}$, $\bm{x}^{\rm max}$ and $\bm{x}^{\rm att}$, we combine them into a compact representation. Practically, we multiply these features with a learnable coefficient $\bm{\gamma} \in \mathcal{R}^{3}$ to control their importance for fusion to obtain $\bm{x}^p \in \mathcal{R}^{T\times C\times 1 \times 1}$ as:

\begin{equation}
  \label{e5}
  \bm{x}^{p} = \bm{x}^{\rm avg} \times \bm{\gamma}_1 + \bm{x}^{\rm max} \times \bm{\gamma}_2 + \bm{x}^{\rm att} \times \bm{\gamma}_3.
\end{equation}

Here, $\bm{\gamma}$ is initialized as a tensor filled with values of $\frac{1}{3}$, and then updated via gradient-based backward propagation in the training process.
Especially, as only one compact representation is used on behalf of the current frame, the computation complexity of calculating correlation maps between adjacent frames can be drastically reduced to only $O(HW)$, in contrast to $O(H^2W^2)$ in CorrNet~\cite{hu2023continuous}. In practice, the computations are notably decreased from 3.64 GFLOPs\footnote{FLOPs denotes the number of multiply-add operations and GFLOPs denotes measuring FLOPs by giga.} to 0.01 GFLOPs, bringing only quite a few extra computations.

Considering sign language is mainly conveyed by consecutive human body motion like hand and arm movements, it's necessary to identify and track the body trajectories in a large temporal neighborhood to understand signs. We strategically enlarge the temporal receptive field of the correlation module to achieve this goal. Specifically, for an input video $\bm{x} \in \mathcal{R}^{T\times C \times H \times W}$, we sample $L$ neighboring frames for each frame to formulate a neighboring frame set $\bm{x}^L \in \mathcal{R}^{T\times C \times L\times H \times W}$. We then recover the last dimension of $\bm{x}^p$ as $\bm{x}^p \in \mathcal{R}^{T\times C\times 1 \times 1\times 1}$ and use it to compute affinities with $\bm{x}^L$ to obtain the local spatial-temporal correlation maps $\bm{A}^L\in \mathcal{R}^{T\times L\times H \times W}$ as:
\begin{align}
  \bm{A}^L = \sum_{i=0}^C \bm{x}^p_{:i}\times  \bm{x}^L_{: i}
\end{align}
where $:$ denotes taking all elements in the corresponding dimension. $L$ can be set as various values in different network stages to capture information of different temporal scales.

Given the spatial-temporal correlation maps $\bm{A}^L$, we constrain values in $\bm{A}^L$ into the range of (0,1) by passing it through a sigmoid function. We further subtract 0.5 from the results to emphasize informative regions with positive values and suppress redundant areas with negative values as:
\begin{equation}
  \label{e7}
  \hat{\bm{A}}^L = {\rm sigmoid}(\bm{A}_L)-0.5.
\end{equation}

After identifying the correlations between adjacent frames, we incorporate them back into each frame to reason about the local human body movements. Specifically, we recover the second dimension of the cross-frame correlations $\hat{\bm{A}}^L$ and repeat it for $C$ times to obtain $\hat{\bm{A}}^L\in \mathcal{R}^{T\times C\times L\times H \times W}$. We then use $\hat{\bm{A}}^L$ to multiply with the features of the neighboring frame set $\bm{x}^L$ to obtain the local human body trajectories $\bm{E}\in \mathcal{R}^{T\times C\times 1\times 1}$ as:
\begin{equation}
  \label{e8}
  \bm{E} = \sum_{l=1}^{L}\sum_{i',j'}{\hat{\bm{A}}^L_{: : l}(i',j') \times \bm{x}^L_{: : l}(i',j')  \times \bm{\beta}_{: : l}}
\end{equation}
where a learnable coefficient $\bm{\beta} \in \mathcal{R}^{1\times 1\times L\times 1\times 1}$ is attached to measure the importance of different neighboring frames. $\bm{\beta}$ is initialized as a tensor filled with values of $\frac{1}{L}$, and updated via gradient-based backward propagation in the training process. This correlation calculation is repeated for each frame in a video to track body trajectories in videos.

\begin{figure}[t]
  \centering
  \includegraphics[width=\linewidth]{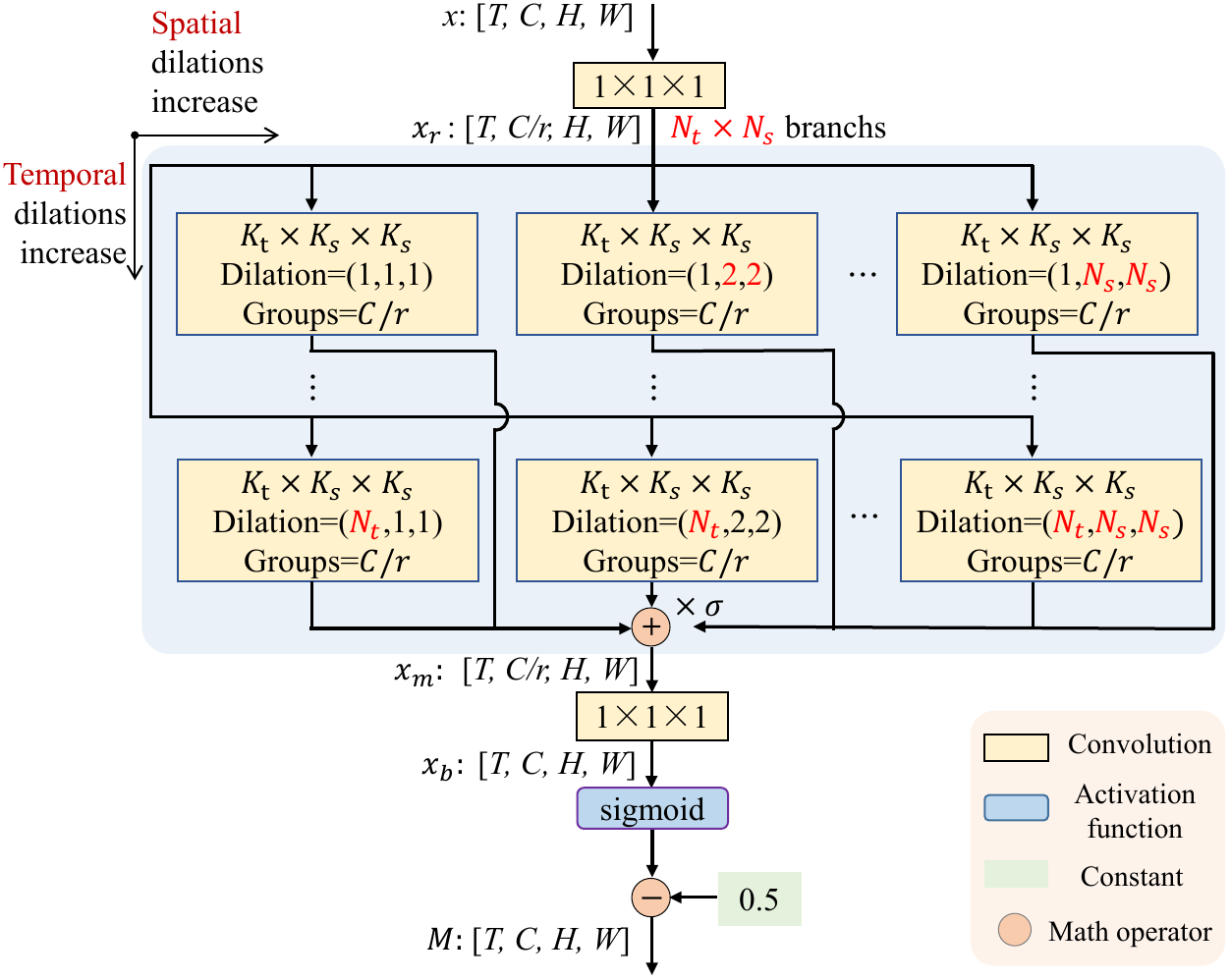}
  \caption{Illustration for our identification module. To avoid heavy computations in identifying informative spatial regions when modeling local spatial-temporal information, we decompose the spatial-temporal modeling structure along the spatial and temporal dimensions simultaneously to form a multiscale architecture, enlarging the model capacity.}
  \label{fig4}
  \end{figure}

\subsection{Identification Module}
The correlation module computes correlation maps among spatial-temporal neighboring patches to model cross-frame interactions. However, not all regions play an equal role in sign expression. Therefore, it's critical to selectively emphasize informative regions that carry essential body trajectories within the current frame $x_t$ and suppress background noise and non-critical elements. To achieve this goal, we present an identification module to dynamically emphasize these informative spatial regions. Specifically, as informative regions like hand and face are misaligned in adjacent frames, the identification module leverages the closely correlated local spatial-temporal features to tackle the misalignment and locate informative spatial regions. 

As shown in Fig.~\ref{fig4}, the identification module first projects input features $\bm{x}\in \mathcal{R}^{T \times C\times H \times W}$ into $\bm{x}_r\in \mathcal{R}^{T \times C/r\times H \times W}$ with a $1\times 1\times 1$ convolution to decrease the computations, with a channel reduction factor $r$ as 16 by default.

As the informative regions, e.g., hands and face, are not exactly aligned in adjacent frames, it's necessary to consider a large spatial-temporal neighborhood to identify these features. Instead of directly employing a large 3D spatial-temporal kernel, we present a multi-scale paradigm by decomposing it into parallel branches of progressive dilation rates to reduce required computations and increase the model capacity. 

Specifically, as shown in Fig.~\ref{fig4}, with a same small base convolution kernel of $K_t \times K_s \times K_s$, we employ multiple convolutions with their dilation rates increasing along spatial and temporal dimensions concurrently. The spatial and temporal dilation rates range within (1, $N_s$) and (1, $N_t$), respectively, resulting in total $N_s\times N_t$ branches. Group-wise convolutions are employed for each branch to reduce parameters and computations. Features from different branches are multiplied with learnable coefficients \{$\bm{\sigma}_1, \dots, \bm{\sigma}_{N_s\times N_t}$\} to control their importance, and then added to mix information from branches of various spatial-temporal receptive fields as:
\begin{equation}
  \label{e9}
  \bm{x}_m = \sum_{i=1}^{N_s}\sum_{j=1}^{N_t}{\bm{\sigma}_{ij} \times {\rm conv}^s_{ij}(\bm{x}_r)}
\end{equation}
where the group-wise convolution ${\rm conv}^s_{i,j}$ of different branches receives features of different spatial-temporal neighborhoods, with dilation rate $(j, i, i)$.

After receiving features from a large spatial-temporal neighborhood, $x_m$ passes through a convolution with kernel size of 1 to project the features into $\bm{x}_b\in \mathcal{R}^{T \times C\times H \times W}$ to recover the channels from $C/r$ to $C$. We then pass $\bm{x}_b$ through a sigmoid function to generate attention maps with values ranging within (0,1), which are further subtracted from 0.5 to obtain $\bm{M}\in \mathcal{R}^{T \times C\times H \times W}$ to emphasize informative regions with positive values and suppress redundant areas with negative values as:
\begin{equation}
  \label{e10}
  \bm{M} = {\rm sigmoid}({\rm conv}_{1\times 1\times 1}(\bm{x}_m)) -0.5.
\end{equation}

Given the attention maps $\bm{M}$ to identify informative regions, it's multiplied with the cross-frame interactions $\bm{E}$ computed by the correlation module to emphasize critical spatial regions that convey body trajectories and suppress others like background or noise. This refined trajectory information is finally incorporated into original spatial features $\bm{x}$ via a residual connection as:
\begin{equation}
\label{e11}
\bm{y} = \bm{x} + \bm{\alpha} \bm{E} \times \bm{M}.
\end{equation}
Here, $\bm{\alpha}$ is initialized as zero to keep the original spatial features and makes the model keep original behaviors.

\begin{figure}[t]
  \centering
  \includegraphics[width=0.75\linewidth]{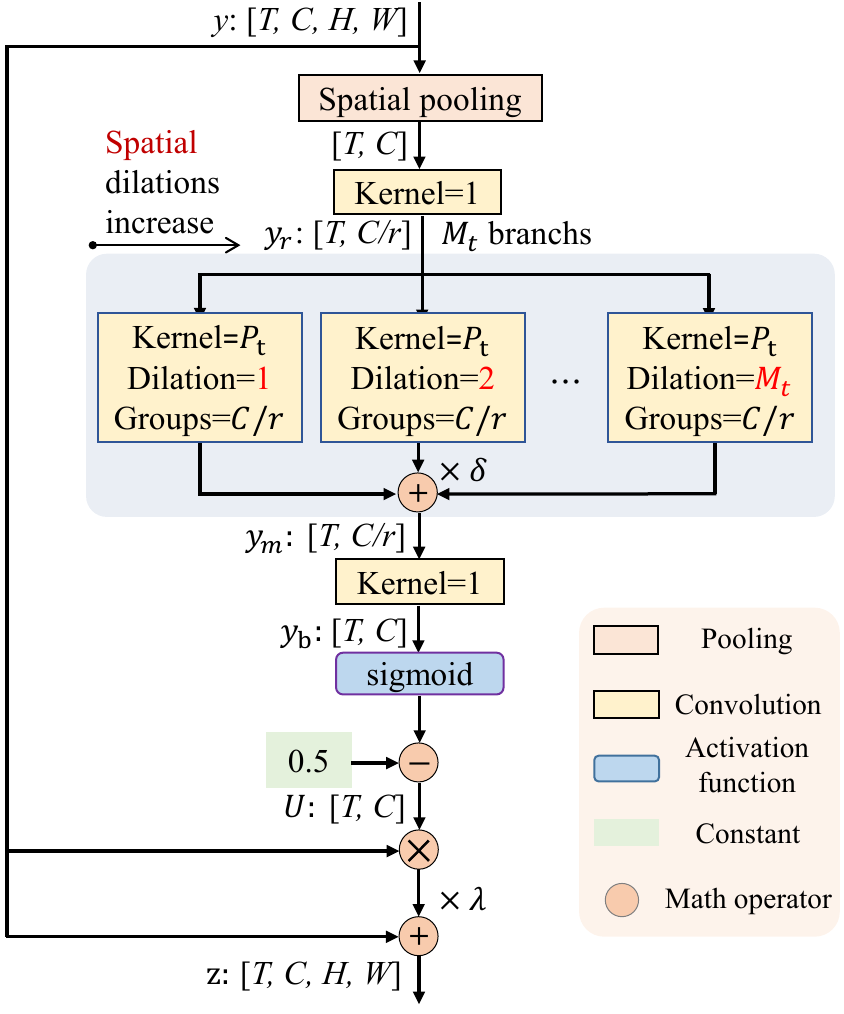}
  \caption{Illustration for our temporal attention module. We employ a temporal multiscale architecture to aggregate local temporal information to dynamically evaluate the contributions of each frame in a lightweight manner.}
  \label{fig5}
  \end{figure}

\subsection{Temporal Attention Module}
The above modules effectively identify the critical cross-frame interactions within informative spatial regions. However, across the entire video, not all frames are equally important in expressing sign language. Some frames carry crucial information while others merely convey idle meanings. To address this, we introduce a temporal attention module. Drawing inspiration from the design principles of the identification module, we dynamically consider the importance of different frames to adaptively emphasize the keyframes and suppress others.

Fig.~\ref{fig5} gives the overview of the temporal attention module. Given the input features $\bm{y}\in \mathcal{R}^{T \times C\times H \times W}$ generated by the correlation module and identification module, we first perform spatial pooling to eliminate the spatial dimensions, and then project the features into $\bm{y}_r\in \mathcal{R}^{T \times C/r\times H \times W}$ with a convolution kernel size of 1 to decrease the computations. 

To sufficiently evaluate the contributions of different frames, we propose a multiscale architecture to leverage the local information in a large temporal neighborhood. In specific, as shown in Fig.~\ref{fig5}, with a same small temporal kernel of $P_t$, multiple parallel depth-wise convolutions are concurrently employed with different dilation rates ranging from 1 to $M_t$ to model information from various temporal receptive fields. Features from different branches are multiplied with learnable coefficients \{$\bm{\delta}_1$, $\dots$, $\bm{\delta}_{P_t}$\} to adjust their importance and added to fuse complementary information from different temporal ranges as:
\begin{equation}
  \label{e12}
  \bm{y}_m = \sum_{i=1}^{P_t}{\bm{\delta}_{i} \times {\rm conv}^t_{i}(\bm{y}_r)}
\end{equation}
where ${\rm conv}^t_{i}$ denotes the group-wise convolution of $i$-th branch with dilation rate $i$.

After receiving the closely correlated spatial-temporal information, $y_m$ passes through a convolution with kernel size of 1 to project the features into $\bm{y}_b\in \mathcal{R}^{T \times C\times H \times W}$ to recover the channels from $C/r$ to $C$. We then pass $\bm{y}_b$ through a sigmoid function to generate temporal attention maps with values ranging within [0,1], which are further subtracted from 0.5 to obtain $\bm{U}\in \mathcal{R}^{T \times C}$ to emphasize keyframes with positive values and suppress others with negative values as:
\begin{equation}
  \label{e13}
  \bm{U} = {\rm sigmoid}({\rm conv}_{1}(\bm{y}_m)) -0.5.
\end{equation}
We then recover the spatial dimensions of $\bm{U}$ to obtain $\bm{U}\in \mathcal{R}^{T \times C \times H\times W}$, and multiply it with input features $\bm{y}$ to dynamically adjust the weights of input frames according to their contributions. These augmented representations are further incorporated into the input features $\bm{y}$ via a residual connection as:
\begin{equation}
  \label{e14}
  \bm{z} = \bm{y} + \bm{\lambda} \bm{y} \times \bm{U}.
  \end{equation}
Here, $\bm{\lambda}$ is initialized as zero during training to avoid hurting the original temporal features.


\section{Experiments}
\subsection{Experimental Setup}
\subsubsection{Datasets.} \textbf{PHOENIX2014}~\cite{koller2015continuous} is recorded from a German weather forecast broadcast with nine actors before a clean background with a resolution of 210 $\times$ 260. It contains 6841 sentences with a vocabulary of 1295 signs, divided into 5672 training samples, 540 development (Dev) samples and 629 testing (Test) samples.

\textbf{PHOENIX2014-T}~\cite{camgoz2018neural} is available for both CSLR and sign language translation tasks. It contains 8247 sentences with a vocabulary of 1085 signs, split into 7096 training instances, 519 development (Dev) instances and 642 testing (Test) instances. It can be used for both \textit{CSLR} and \textit{SLT} tasks.

\textbf{CSL-Daily}~\cite{zhou2021improving} revolves the daily life, recorded indoor at 30fps by 10 signers. It contains 20654 sentences, divided into 18401 training samples, 1077 development (Dev) samples and 1176 testing (Test) samples.  It can be used for both \textit{CSLR} and \textit{SLT} tasks.

\textbf{CSL}~\cite{huang2018video} is collected in the laboratory by fifty signers with a vocabulary size of 178 with 100 sentences. It contains 25000 videos, divided into training and testing sets by a ratio of 8:2.

\subsubsection{Training details.} For fair comparisons, we follow the same setting as state-of-the-art methods~\cite{Min_2021_ICCV,zuo2022c2slr} to prepare our model. We adopt ResNet18~\cite{he2016deep} as the 2D CNN backbone with ImageNet~\cite{deng2009imagenet} pretrained weights. The 1D CNN of state-of-the-art methods is set as a sequence of \{K5, P2, K5, P2\} layers where K$\theta$ and P$\theta$ denotes a 1D convolutional layer and a pooling layer with kernel size of $\theta$, respectively. A two-layer BiLSTM with hidden size 1024 is attached for long-term temporal modeling, followed by a fully connected layer for sentence prediction. We train our models for 80 epochs with initial learning rate 0.001 which is divided by 5 at epoch 40 and 60. Adam~\cite{kingma2014adam} optimizer is adopted as default with weight decay 0.0001 and batch size 2. All input frames are first resized to 256$\times$256, and then randomly cropped to 224$\times$224 with 50\% horizontal flipping and 20\% temporal rescaling during training. During inference, a 224$\times$224 center crop is simply adopted. Following VAC~\cite{Min_2021_ICCV}, we employ the VE loss and VA loss for visual supervision, with weights 1.0 and 25.0, respectively. We adopt the TLP loss~\cite{hu2022temporal} to extract more powerful representations. Our model is trained and evaluated upon a 3090 graphical card. For the SLT task, the translation network is instantiated as a mBART model~\cite{liu2020multilingual}. In practice, we found that the gloss labels are beneficial for SLT. Thus we let the translation process additionally supervised with the recognition loss $\mathcal{L}_{\rm CTC}$, whose final losses can be expressed as: $\mathcal{L}_{\rm T} = \mathcal{L}_{\rm CTC} + \mathcal{L}_{\rm CE}$. We set the learning rate of the visual mapper and translation network as 0.0002 and 1e-6, respectively. We train our models for 40 epochs with learning rates divided by 5 at epoch 20 and 30.

\begin{table}[t]   
  \centering
  \caption{Ablations for the effectiveness of the proposed correlation module, identification module and temporal attention module on the PHOENIX2014 dataset.} 
  \label{tab1} 
  \begin{tabular}{ccccc}
  \toprule
  \midrule
  Correlation & Identification & \makecell{Temporal\\ Weighting} & Dev(\%) & Test(\%)\\
  \midrule
  \ding{56} & \ding{56} & \ding{56}  & 20.2 & 21.0 \\
  \Checkmark & \ding{56} & \ding{56}    & 19.2 & 19.7 \\
  \ding{56} & \Checkmark & \ding{56}   & 19.5 & 20.1\\
  \ding{56} & \ding{56} & \Checkmark  & 19.6 & 20.2\\
  \midrule
  \Checkmark & \Checkmark  & \ding{56}  & 18.4 & 18.7 \\
  \Checkmark & \ding{56} & \Checkmark  & 18.8 & 19.2\\
  \ding{56} & \Checkmark & \Checkmark  & 19.0 & 19.3\\
  \Checkmark & \Checkmark  & \Checkmark  & \textbf{18.0} & \textbf{18.2} \\
  \bottomrule
  \end{tabular}
  \end{table}

\subsubsection{Evaluation Metric.} For the \textbf{CSLR} task, we use Word Error Rate (WER) as the evaluation metric, which is defined as the minimal summation of the \textbf{sub}stitution, \textbf{ins}ertion, and \textbf{del}etion operations to convert the predicted sentence to the reference sentence, as:
\begin{equation}
\label{e11}
\rm WER = \frac{ \#sub+\#ins+\#del}{\#reference}.
\end{equation}
Note that the \textbf{lower} WER, the \textbf{better} accuracy.

For the \textbf{SLT} task, following previous studies~\cite{chen2022two,zhou2021improving}, we use commonly-used metrics in machine translation, including tokenized BLEU~\cite{papineni2002bleu} with ngrams from 1 to 4 (BLEU@1-BLEU@4) and Rouge-L F1 (Rouge)~\cite{lin2004rouge} to evaluate the performance of SLT. The higher value, the better performance. 

\subsection{Ablation Study}
We report ablative results on both development (Dev) and testing (Test) sets of PHOENIX2014 dataset to test the effectiveness of each component in our CorrNet+.

\begin{table}[t]   
  \centering
  \caption{Ablations for the locations of CorrNet+ on the PHOENIX2014 dataset.} 
  \label{tab2} 
  \begin{tabular}{ccccc}
  \toprule
  Stage 2 & Stage 3 & Stage 4 & Dev(\%) & Test(\%)\\
  \midrule
  \ding{56} & \ding{56}  & \ding{56}    & 20.2 & 21.0 \\
    \Checkmark & \ding{56}  & \ding{56}   & 19.3 & 19.9 \\
    \ding{56} & \Checkmark  & \ding{56}   & 19.2 & 19.7\\
    \ding{56} & \ding{56}  & \Checkmark   & 19.0 & 19.5 \\
    \midrule
    \Checkmark & \Checkmark  & \ding{56}   & 18.5 & 18.8 \\
    \Checkmark & \Checkmark  & \Checkmark   & \textbf{18.0} & \textbf{18.2} \\
  \bottomrule
  \end{tabular}
  \end{table}

\begin{table}[t]   
  \centering
  \caption{Ablations for the effectiveness of correlation module on the PHOENIX2014 dataset.} 
  \label{tab4} 
  \begin{tabular}{lccc}
  \toprule
  Configurations  & Dev(\%) & Test(\%) & \makecell{Extra GFLOPs/ \\Original GFLOPs}\\
  \midrule
  CorrNet~\cite{hu2023continuous} & 18.8 & 19.4 & 3.600 / 3.640 \\
  CorrNet+ & \textbf{18.0} & \textbf{18.2} & \textbf{0.010 / 3.640}\\
  \midrule
  - & 20.2 & 21.0 & -\\
  $L$=[2,2,2] & 19.0  & 19.0 & 0.007 / 3.640 \\
  $L$=[6,6,6] & 18.5 & 18.8 & 0.010 / 3.640 \\
  $L$=[10,10,10] & 18.4 & 18.7 & 0.012 / 3.640 \\
  $L$=[2,6,10] & \textbf{18.0} & \textbf{18.2} & \textbf{0.010 / 3.640} \\
  $L$=[10,6,2] & 18.6 & 18.8 & 0.012 / 3.640\\
  $L$=[6,10,14] & 18.3 & 18.4 & 0.012 / 3.640\\
  \bottomrule
  \end{tabular}
  \end{table}
\textbf{Effectiveness of the proposed modules.} Tab.~\ref{tab1} provides a comprehensive analysis of the effectiveness of the proposed modules. We notice that using any of the proposed three modules yields a notable accuracy boost, with 19.2\% \& 19.7\%, 19.5\% \& 20.1\% accuracy and 19.6 \& 20.2\% WER on the Dev and Test Sets, respectively. Notably, the correlation module offers the most substantial accuracy improvement. Combining any two modules further activates the effectiveness with 18.4\% \& 18.7\%, 18.8\% \& 19.2\% and 19.0\% \& 19.3\% WER on the Dev and Test Sets, respectively. We notice that combining the correlation module and the identification module gives the most performance promotion. When employing all proposed modules, the accuracy reaches the peak with absolute 18.0\% \& 18.2\% WER, giving +2.2\% \& +2.8\% accuracy boost.

\textbf{Effects of locations for CorrNet+.} Tab~\ref{tab2} ablates the locations of our proposed modules in Stage 2, 3 or 4. We observe that choosing any one of these locations brings a notable accuracy boost, with 19.3\% \& 19.9\%, 19.2\% \& 19.7\% and 19.0\% \& 19.5\% WER. When combining two or more locations, a larger accuracy gain is witnessed. The accuracy reaches the peak when proposed modules are placed after Stage 2, 3 and 4, with 18.0\% \& 18.2\% accuracy, which is adopted by default.

\begin{table}[t]   
  \centering
  \caption{Ablations for the effectiveness of the aggregation functions in correlation module on the PHOENIX2014 dataset.} 
  \label{tab5} 
  \begin{tabular}{ccc}
  \toprule
  Aggregation function  & Dev(\%) & Test(\%)\\
  \midrule
  AvgPool  & 18.5 & 18.8 \\
  AvgPool \& MaxPool & 18.3 & 18.5 \\
  AvgPool \& MaxPool \& AttPool & \textbf{18.0} & \textbf{18.2} \\
  \bottomrule
  \end{tabular}
  \end{table}

\begin{table}[t]   
  \centering
  \caption{Ablations for the multi-scale architecture of identification module on the PHOENIX14 dataset.} 
  \label{tab6} 
  \begin{tabular}{cccc}
  \hline
  Configuration & Dev(\%) & Test(\%)\\
  \hline
  - & 20.2 & 21.0\\
  \hline
  $N_t$=4, $N_s$=\textbf{1}  & 18.8  & 18.9\\
  $N_t$=4, $N_s$=\textbf{2} & 18.4 & 18.6 \\
  $N_t$=4, $N_s$=\textbf{3} & \textbf{18.0} & \textbf{18.2} \\
  $N_t$=4, $N_s$=\textbf{4}  & 18.3  &18.5 \\
  \hline
  $N_t$=\textbf{2}, $N_s$=3 & 18.6  & 18.7 \\
  $N_t$=\textbf{3}, $N_s$=3 & 18.3  & 18.5 \\
  $N_t$=\textbf{4}, $N_s$=3 & \textbf{18.0} & \textbf{18.2} \\
  $N_t$=\textbf{5}, $N_s$=3  & 18.5 & 18.6 \\
  \hline
  $K_t$=\textbf{9}, $K_s$=\textbf{7} & 19.1 & 19.2 \\
  \hline
  \end{tabular}
  \vspace{-5px}
  \end{table}
\textbf{Study of the effectiveness of correlation module.} In the upper part of Tab.~\ref{tab4}, we first verify the effectiveness of CorrNet+ by comparing it to the CorrNet~\cite{hu2023continuous}. By computing correlation maps between all spatial patches among consecutive frames, CorrNet promotes the WER to 18.8\% \& 19.4\% on the Dev and Test sets, respectively. However, it raises substantial computational overhead (3.60 GFLOPs), nearly equivalent to the entire model’s computation (3.64 GFLOPs). Instead, by compressing the features of each frame, CorrNet+ notably decreases the incurred computations from 3.60 GFLOPs to 0.01 GFLOPs and brings +0.8\% \& +1.2\% accuracy boost, achieving a better accuracy-computation trade-off. In the lower part of Tab.~\ref{tab4}, we investigate the effects of the temporal receptive filed $L=\{L_1, L_2, L_3\}$ across three network stages for the correlation module. When disabling $L$, the model degenerates into our baseline. We observe that when setting $L=[2,2,2]$ (focusing solely on adjacent frames) CorrNet+ outperforms the baseline by 1.2\% \& 2.0\% on the Dev and Test sets, respectively. Gradually increasing $L$ from [1,1,1] to [5,5,5] consistently improves accuracy with similar computational costs. We then investigate different configurations for the temporal receptive fields as network stages progress. We notice that $L=[2,6,10]$ yields the peak accuracy, and either reversing the order of $L$ of further increasing $L$ would degrade the performance. 

\textbf{Study on the effectiveness of aggregation functions in correlation module.} We verify the effectiveness of the aggregation functions for the correlation module in Tab.~\ref{tab5}. It's observed that by solely using the average aggregation function, CorrNet+ already achieve better results (18.5\% \& 18.8\%) than CorrNet (18.8\% \& 19.2\%). When incorporating both the maximum and attention aggregation functions, the performance is further promoted to 18.3\% \& 18.5\% and 18.0\% \& 18.2\%, underscoring the complementarity of the proposed aggregation functions.

\begin{table}[t]   
  \centering
  \caption{Ablations for the configurations of temporal attention module on the PHOENIX2014 dataset.} 
  \label{tab7} 
  \begin{tabular}{cccc}
  \toprule
  Configuration & Dev(\%) & Test(\%)\\
  \midrule
  - & 20.2 & 21.0\\
  \midrule
  $M_t$=1 & 18.6  & 18.7 \\
  $M_t$=2 & 18.3  & 18.5 \\
  $M_t$=3 & \textbf{18.0} & \textbf{18.2} \\
  $M_t$=4  & 18.2 & 18.4 \\
  $M_t$=5  & 18.3 & 18.5\\
  $P_t$=5 & 19.1 & 19.2 \\
  \midrule
  $\bm{U} \odot \bm{y}$  & 21.2  & 22.1 \\
  $\bm{U} \odot \bm{y} + \bm{y}$ & 19.6 & 20.3 \\
  $(\bm{U} -0.5) \odot \bm{y}$ & 18.5  & 18.8 \\
  $(\bm{U} -0.5) \odot \bm{y} +\bm{y} $ & \textbf{18.0}  & \textbf{18.2}\\
  \bottomrule
  \end{tabular}
  \end{table}

\begin{table}[t]   
  \centering
  \caption{Ablations for the generalizability of CorrNet over multiple backbones on the PHOENIX2014 dataset.} 
  \label{tab8} 
  \begin{tabular}{lcc}
    \toprule
    Configuration & Dev(\%) & Test(\%)\\
    \midrule
    SqueezeNet~\cite{hu2018squeeze}  & 22.2 & 22.6 \\
    \quad w/ CorrNet+  & \textbf{19.4}  & \textbf{19.6} \\
    \midrule
    ShuffleNet V2~\cite{ma2018shufflenet} & 21.7 & 22.2 \\
    \quad w/ CorrNet+  & \textbf{19.1}  & \textbf{19.5} \\
    \midrule
    GoogleNet~\cite{szegedy2015going}  & 21.4 & 21.5\\
    \quad w/ CorrNet+  & \textbf{18.9} & \textbf{19.0} \\
    \midrule
    RegNetX-800mf~\cite{radosavovic2020designing}  & 20.4 & 21.2 \\
    \quad w/ CorrNet+  & \textbf{18.3}  & \textbf{18.4} \\
    \midrule
    RegNetY-800mf~\cite{radosavovic2020designing} & 20.1 & 20.8\\
    \quad w/ CorrNet+  & \textbf{17.8}  & \textbf{18.0} \\
    \bottomrule
    \end{tabular}
    \vspace{-5px}
    \end{table}

\begin{table}[t]   
  \centering
  \setlength\tabcolsep{3pt}
  \caption{Comparison with other methods of spatial-temporal attention or temporal reasoning on the PHOENIX2014 dataset.} 
  \label{tab9} 
  \begin{tabular}{lcc}
  \toprule
  Method & Dev(\%) & Test(\%)\\
  \midrule
  - & 20.2 & 21.0\\
  w/ SENet~\cite{hu2018squeeze}  & 19.8  & 20.4 \\
  w/ CBAM~\cite{woo2018cbam} & 19.7 & 20.2 \\
  w/ NLNet~\cite{wang2018non} & - & -\\
  \midrule
  I3D~\cite{carreira2017quo} & 22.6  & 22.9 \\
  R(2+1)D~\cite{tran2018closer}  & 22.4  & 22.3 \\
  TSM~\cite{lin2019tsm} & 19.9 & 20.5 \\
  \midrule
  CorrNet+ & \textbf{18.0} & \textbf{18.2} \\
  \bottomrule
  \end{tabular}
  \end{table}

\begin{table}[t]   
  \centering
  \setlength\tabcolsep{3pt}
  \caption{Comparison with other methods that explicitly exploit hand and face features on the PHOENIX2014 dataset.} 
  \label{tab10} 
  \begin{tabular}{lcc}
  \toprule
  Method & Dev(\%) & Test(\%)\\
  \midrule
  CNN+HMM+LSTM~\cite{koller2019weakly} & 26.0  & 26.0 \\
  DNF~\cite{cui2019deep} & 23.1 & 22.9\\
  STMC~\cite{zhou2020spatial} & 21.1  & 20.7 \\
  C$^2$SLR~\cite{zuo2022c2slr} & 20.5  & 20.4 \\
  \midrule
  CorrNet+ & \textbf{18.0} & \textbf{18.2} \\
  \bottomrule
  \end{tabular}
  \end{table}
\textbf{Study on the multi-scale architecture of identification module.} In Tab.~\ref{tab6}, without identification module, our baseline achieves 20.2\% and 21.0\% WER on the Dev and Test Set, respectively. The base kernel size is set as $3\times 3\times 3$ for $K_t\times K_s\times K_s$. When fixing $N_t$=4 and varying spatial dilation rates to expand spatial receptive fields, a larger $N_s$ consistently brings better accuracy. When $N_s$ reaches 3, it brings no more accuracy gain. Consequently, we set $N_s$ as 3 by default and investigate the impact of $N_t$. Notably, increasing $N_t$ to 5 or decreasing $N_t$ to 2 and 3 achieves worse accuracy. We thus adopt $N_t$ as 4 by default. We also compare our proposed multi-scale architecture with a normal implementation of more parameters. The receptive field of the identification module with $N_t$=4, $N_s$=3 is identical to a normal convolution with $K_t$=9 and $K_s$=7. As shown in the bottom of Tab.~\ref{tab6}, although a normal convolution owns more parameters and computations than ours, it performs worse than our method which verifies the effectiveness of our proposed architecture.

\textbf{Study on the configurations of temporal attention module.} In the upper part of Tab.~\ref{tab7}, we investigate the effects for the number of branches $M_t$ in the temporal attention module. We notice that as $M_t$ increases, the performance consistently rises until it reaches 3, and a larger $M_t$ can't bring more performance gain. We thus set $M_t=3$ by default. We then investigate the efficacy of the multiscale architecture by comparing it against a large convolution with kernel size $P_t$ of 5, which has the same temporal receptive field. We observe that our design outperforms it by a large margin with lower computational costs. In the lower part of Tab.~\ref{tab7}, we explore the implementations of the temporal attention module to augment original features. Initially, a direct multiplication of the attention maps $\bm{U}$ with input features $y$ severely degrades performance due to the disruption of input feature distributions. However, when implemented residually by adding y, the expression $\bm{U} \odot \bm{y} +\bm{y} $ notably mitigates this phenomenon, resulting in performance gains of +0.6\% and +0.7\% on the Dev and Test Sets, respectively. We further subtract 0.5 from the attention maps $\bm{U}$ to emphasize or suppress certain positions, and then element-wisely multiply it with $\bm{y}$. This refined implementation brings +1.1\% \& +1.5\% performance boost. Finally, we update this implementation in a residual way by adding input features $\bm{y}$ as $(\bm{U} -0.5) \odot \bm{y} +\bm{y} $, achieving a notable performance boost by +2.2\% \& +2.8\%.

\textbf{Generalizability of CorrNet+.} We deploy CorrNet+ upon multiple backbones, including SqueezeNet~\cite{hu2018squeeze}, ShuffleNet V2~\cite{ma2018shufflenet}, GoogLeNet~\cite{szegedy2015going}, RegNetX-800mf~\cite{radosavovic2020designing} and RegNetY-800mf~\cite{radosavovic2020designing} to validate its generalizability in Tab.~\ref{tab8}. It's observed that our proposed model generalizes well upon different backbones, bringing +2.8\% \& +3.0\%, +2.6\% \& +2.7\%, +2.5\% \& +2.5\%, +2.1\% \& +2.8\% and +2.3\% \& +2.8\% accuracy boost on the Dev and Test Sets, respectively.

\textbf{Comparisons with other spatial-temporal reasoning methods.} 
Tab.~\ref{tab9} compares our approach with other methods of spatial-temporal reasoning ability. SENet~\cite{hu2018squeeze} and CBAM~\cite{woo2018cbam} perform channel attention to emphasize key information. NLNet~\cite{wang2018non} employs non-local means to aggregate spatial-temporal information from other frames. I3D~\cite{carreira2017quo} and R(2+1)D~\cite{tran2018closer} deploys 3D or 2D+1D convolutions to capture spatial-temporal features. TSM~\cite{lin2019tsm} adopts temporal shift operation to obtain features from adjacent frames. In the upper part of Tab.~\ref{tab9}, one can see CorrNet+ largely outperforms other attention-based methods, i.e., SENet, CBAM and NLNet, for its superior ability to identify and aggregate body trajectories. NLNet is out of memory due to its quadratic computational complexity with spatial-temporal size. In the bottom part of Tab.~\ref{tab9}, we observed that I3D and R(2+1)D demonstrate degraded accuracy, which may be attributed to their limited spatial-temporal receptive fields and increased training complexity. TSM slightly brings 0.3\% \& 0.3\% accuracy boost. Our proposed approach significantly outperforms these methods, affirming its efficacy in aggregating salient spatial-temporal information from even distant spatial neighbors.

\begin{table}[t]   
  \centering
  \setlength\tabcolsep{4pt}
  \caption{Comparison with state-of-the-art methods on the PHOENIX2014 and PHOENIX2014-T datasets over the CSLR setting. $*$ indicates extra clues such as face or hand features are included by additional networks or pre-extracted heatmaps.} 
  \label{tab11}
  \begin{tabular}{lccccccc}
  \toprule
  \multirow{3}{*}{Method} & \multicolumn{4}{c}{PHOENIX2014} & \multicolumn{2}{c}{PHOENIX2014-T} \\
  & \multicolumn{2}{c}{Dev(\%)} & \multicolumn{2}{c}{Test(\%)} &  \multirow{2}{*}{Dev(\%)} & \multirow{2}{*}{Test(\%)}\\
  & del/ins & WER & del/ins& WER & & \\
  \midrule
  SFL~\cite{niu2020stochastic} & 7.9/6.5 & 26.2 & 7.5/6.3& 26.8 & 25.1&26.1\\
  FCN~\cite{cheng2020fully} & - & 23.7 & -& 23.9 & 23.3& 25.1\\
  CMA~\cite{pu2020boosting} & 7.3/2.7 & 21.3 & 7.3/2.4 & 21.9  & -&-\\
  VAC~\cite{Min_2021_ICCV} & 7.9/2.5 & 21.2 &8.4/2.6 & 22.3 &- &-\\
  SMKD~\cite{hao2021self}&6.8/2.5 &20.8 &6.3/2.3 & 21.0 & 20.8 & 22.4\\
  CVT-SLR~\cite{zheng2023cvt} & 6.4/2.6 & 19.8 & 6.1/2.3 & 20.1 & 19.4 & 20.3\\
  TLP~\cite{hu2022temporal}& 6.3/2.8 & 19.7 & 6.1/2.9 & 20.8 & 19.4  & 21.2 \\
  CoSign-2s~\cite{jiao2023cosign}& - & 19.7 & - &20.1 & 19.5 & 20.1\\
  AdaSize~\cite{hu2024scalable} & 7.0/2.6 & 19.7 & 7.2/3.1 & 20.9 & 19.7 & 21.2\\
  AdaBrowse+~\cite{hu2023adabrowse} & 6.0/2.5 & 19.6 & 5.9/2.6 & 20.7 & 19.5 & 20.6\\
  SEN~\cite{hu2023self} & 5.8/2.6 &  19.5 &  7.3/4.0 &  21.0 &  19.3 &  20.7 \\
  CTCA~\cite{guo2023distilling} & 6.2/2.9 & 19.5 & 6.1/2.6 & 20.1 & 19.3 & 20.3\\
  RadialCTC~\cite{min2022deep} & 6.5/2.7 & 19.4 & 6.1/2.6 & 20.2 & - & -\\
  \midrule
  SLT$^*$~\cite{camgoz2018neural}  & - & - & - & - & 24.5 & 24.6\\
  C+L+H$^*$~\cite{koller2019weakly}  & - &26.0 & - & 26.0 & 22.1 & 24.1 \\
  DNF$^*$~\cite{cui2019deep}  & 7.3/3.3 &23.1& 6.7/3.3 & 22.9 & - & -\\
  STMC$^*$~\cite{zhou2020spatial}& 7.7/3.4 &21.1 & 7.4/2.6 & 20.7 & 19.6 & 21.0\\
  C$^2$SLR$^*$~\cite{zuo2022c2slr} & - & 20.5 &- & 20.4 & 20.2 & 20.4  \\
  \midrule
  \textbf{CorrNet+} & 5.3/2.7 & \textbf{18.0} & 5.6/2.4 & \textbf{18.2} & \textbf{17.2} & \textbf{19.1} \\

  \bottomrule   
  \end{tabular}  
  \vspace{-15px}
\end{table}

\begin{minipage}[!t]{\textwidth}
  \begin{minipage}[t]{0.26\textwidth}
  \footnotesize
  \makeatletter\def\@captype{table}
  \setlength\tabcolsep{4pt}
  \caption{Comparison with state-of-the-art methods on the CSL-Daily dataset~\cite{zhou2021improving} over the CSLR setting.} 
  \label{tab12}
    \begin{tabular}{lccc}
    \toprule
    Method&  Dev(\%) & Test(\%)\\
    \midrule
    BN-TIN~\cite{zhou2021improving} & 33.6  & 33.1 \\
    FCN~\cite{cheng2020fully} & 33.2  & 32.5 \\
    Joint-SLRT~\cite{camgoz2020sign}  & 33.1  & 32.0 \\
    TIN-Iterative~\cite{cui2019deep}  & 32.8  & 32.4\\
    CTCA~\cite{guo2023distilling} & 31.3 & 29.4 \\
    AdaSize~\cite{hu2024scalable} & 31.3 & 30.9\\
    AdaBrowse+~\cite{hu2023adabrowse} & 31.2  & 30.7 \\
    SEN~\cite{hu2023self} & 31.1 & 30.7 \\
    \midrule
    \textbf{CorrNet+} & \textbf{28.6} & \textbf{28.2} \\
    \bottomrule
    \end{tabular}  
  \end{minipage}
  \begin{minipage}[t]{0.20\textwidth}
  \footnotesize   
  \makeatletter\def\@captype{table}
  \setlength\tabcolsep{4pt}
  \caption{Comparison with state-of-the-art methods on the CSL dataset~\cite{huang2018video} over the CSLR setting.} 
  \label{tab13}
  \begin{tabular}{lc}
    \toprule
    Method&  WER(\%)\\
    \midrule
    LS-HAN~\cite{huang2018video}  & 17.3 \\
    SubUNet~\cite{cihan2017subunets}   & 11.0\\
    SF-Net~\cite{yang2019sf} & 3.8 \\
    FCN~\cite{cheng2020fully}   & 3.0 \\
    STMC~\cite{zhou2020spatial}  & 2.1 \\
    VAC~\cite{Min_2021_ICCV} & 1.6 \\
    C$^2$SLR~\cite{zuo2022c2slr} & 0.9 \\
    SEN~\cite{hu2023self} & 0.8\\
    \midrule
    \textbf{CorrNet+} & \textbf{0.7} \\
    \bottomrule
    \end{tabular}  
  \end{minipage}
  \vspace{10pt}
\end{minipage}

\textbf{Comparisons with previous methods equipped with hand or face features.} Many previous CSLR methods explicitly leverage hand and face features for better recognition by employing multiple input streams~\cite{koller2019weakly}, human body keypoints~\cite{zhou2020spatial,zuo2022c2slr} and pre-extracted hand patches~\cite{cui2019deep}. They require extra resource-intensive pose-estimation networks like HRNet~\cite{wang2020deep} or additional multiple training stages. Our approach doesn't rely on extra supervision and could be end-to-end trained to dynamically attend to body trajectories like hand and face actions in a self-motivated way. Tab.~\ref{tab10} shows that our method could outperform these methods by a large margin with much fewer computations.
\begin{table*}[t]   
  \centering
  \setlength\tabcolsep{3pt}
  \caption{Comparison with state-of-the-art methods on the PHOENIX2014-T dataset~\cite{camgoz2018neural} and CSL-Daily dataset~\cite{zhou2021improving} over the SLT setting.} 
    \label{tab14}
  \begin{tabular}{cl|ccccc|ccccc}
    \hline 
    \hline
    \multicolumn{12}{c}{ PHOENIX2014-T } \\
    \hline 
    & \multirow{2}{*}{Method}& \multicolumn{5}{c|}{ Dev } & \multicolumn{5}{c}{ Test }  \\
    & &Rouge & BLEU1 & BLEU2 & BLEU3 & BLEU4 & Rouge & BLEU1 &BLEU2 & BLEU3 & BLEU4 \\
    \hline 
    \multirow{6}{*}{ Sign2Gloss2Text } & SL-Luong~\cite{camgoz2018neural} & 44.14 & 42.88 & 30.30 & 23.02 & 18.40 & 43.80 & 43.29 & 30.39 & 22.82 & 18.13 \\
    & SignBT~\cite{zhou2021improving} & 49.53 & 49.33 & 36.43 & 28.66 & 23.51 & 49.35 & 48.55 & 36.13 & 28.47 & 23.51 \\
    & STMC-Transf~\cite{yin2020better} & 46.31 & 48.27 & 35.20 & 27.47 & 22.47 & 46.77 & 48.73 & 36.53 & 29.03 & 24.00 \\
    & MMTLB~\cite{chen2022simple}  & 50.23 & 50.36 & 37.50 & 29.69 & 24.63 & 49.59 & 49.94 & 37.28 & 29.67 & 24.60 \\
    & TwoStream-SLT~\cite{chen2022two} & 52.01 & 52.35 & 39.76 & 31.85 & 26.47 & 51.59 & 52.11 & 39.81 & 32.00 & 26.71 \\
    & SLTUNET~\cite{zhang2023sltunet} & 49.61 & - & - & - & 25.36 & 49.98 & 50.42 & 39.24 & 31.41 & 26.00\\
    \hline
    \multirow{6}{*}{ Sign2Text }&  SL-Luong~\cite{camgoz2018neural} & 31.80 & 31.87 & 19.11 & 13.16 & 9.94 & 31.80 & 32.24 & 19.03 & 12.83 & 9.58 \\
    & Joint-SLRT~\cite{camgoz2020sign} &- & 47.26 & 34.40 & 27.05 & 22.38 & - & 46.61 & 33.73 & 26.19 & 21.32 \\
    & STMC-T~\cite{zhou2021spatial} & 48.24 & 47.60 & 36.43 & 29.18 & 24.09 & 46.65 & 46.98 & 36.09 & 28.70 & 23.65 \\
    & SignBT~\cite{zhou2021improving}& 50.29 & 51.11 & 37.90 & 29.80 & 24.45 & 49.54 & 50.80 & 37.75 & 29.72 & 24.32 \\
    & MMTLB~\cite{chen2022simple}& 53.10 & 53.95 & 41.12 & 33.14 & 27.61 & 52.65 & 53.97 & 41.75 & 33.84 & 28.39 \\
    & SLTUNET~\cite{zhang2023sltunet} & 52.23 & - & - & - & 27.87 & 52.11 & 52.92 & 41.76 & 33.99 & 28.47\\
    & TwoStream-SLT~\cite{chen2022two} & 54.08 & 54.32 & 41.99 & 34.15 & 28.66 & 53.48 & 54.90 & 42.43 & 34.46 & 28.95 \\
    & \textbf{CorrNet+} & \textbf{54.54} & \textbf{54.56} & \textbf{42.31} & \textbf{34.48} & \textbf{29.13} & \textbf{53.76} & \textbf{55.32} & \textbf{42.74} & \textbf{34.86} & \textbf{29.42}\\
    \hline
    \hline
    \multicolumn{12}{c}{ CSL-Daily } \\
    \hline 
    & \multirow{2}{*}{Method}& \multicolumn{5}{c|}{ Dev } & \multicolumn{5}{c}{ Test }  \\
    & &Rouge & BLEU1 & BLEU2 & BLEU3 & BLEU4 & Rouge & BLEU1 &BLEU2 & BLEU3 & BLEU4 \\
    \hline 
    \multirow{6}{*}{ Sign2Gloss2Text } & SL-Luong~\cite{camgoz2018neural} & 40.18 & 41.46 & 25.71 & 16.57 & 11.06 & 40.05 & 41.55 & 2573 & 16.54 & 11.03 \\
    & SignBT~\cite{zhou2021improving} & 48.38 & 50.97 & 36.16 & 26.26 & 19.53 & 48.21 & 50.68 & 36.00 & 26.20 & 19.67 \\
    & MMTLB~\cite{chen2022simple}  & 51.35 & 50.89 & 37.96 & 28.53 & 21.88 & 51.43 & 50.33 & 37.44 & 28.08 & 21.46 \\
    & SLTUNET~\cite{zhang2023sltunet} & 52.89 & - & - & - & 22.95 & 53.10 & 54.39 &  40.28 & 30.52 & 23.76 \\
    & TwoStream-SLT~\cite{chen2022two} & 53.91 & 53.58 & 40.49 & 30.67 & 23.71 & 54.92 & 54.08 & 41.02 & 31.18 & 24.13 \\
    \hline
    \multirow{6}{*}{ Sign2Text }&  SL-Luong~\cite{camgoz2018neural} & 34.28 & 34.22 & 19.72 & 12.24 & 7.96 & 34.54 & 34.16 & 19.57 & 11.84 & 7.56 \\
    & SignBT~\cite{zhou2021improving} & 49.49& 51.46 & 37.23 & 27.51 & 20.80 & 49.31 & 51.42 & 37.26 & 27.76 & 21.34 \\
    & MMTLB~\cite{chen2022simple} & 53.38 & 53.81 & 40.84 & 31.29 & 24.42 & 53.25 & 53.31 & 40.41 & 30.87 & 23.92 \\
    & SLTUNET~\cite{zhang2023sltunet} & 53.58 & - & -& - &23.99 & 54.08 & 54.98 & 41.44 & 31.84 & 25.01 \\
    & TwoStream-SLT~\cite{chen2022two} & 55.10 & 55.21 & 42.31 & 32.71 & 25.76 & 55.72 & 55.44 & 42.59 & 32.87 & 25.79 \\
    & \textbf{CorrNet+} & \textbf{55.52} & \textbf{55.64} & \textbf{42.78} & \textbf{33.13} & \textbf{26.14} & \textbf{55.84} & \textbf{55.82} & \textbf{42.96} & \textbf{33.26} & \textbf{26.14}\\
    \hline
    \end{tabular}
  \end{table*}  

\begin{figure*}[!h]
  \centering
  \includegraphics[width=0.9\linewidth]{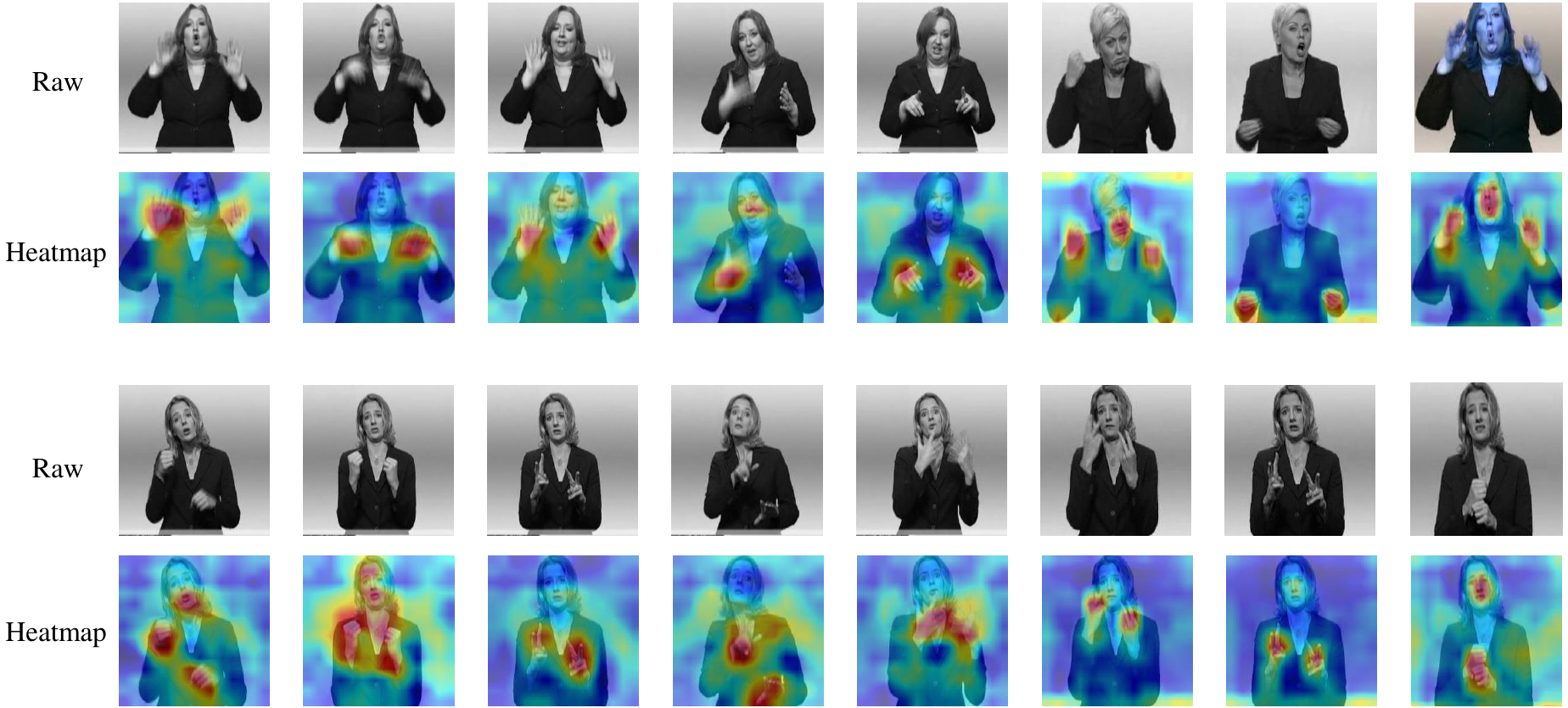}
  \caption{Visualizations of heatmaps by Grad-CAM~\cite{selvaraju2017grad}. Top: raw frames; Bottom: heatmaps of identification module. Our identification module could generally focus on the human body (light yellow areas) and especially pays attention to informative regions like hands and face (dark red areas) to track body trajectories.}
  \label{fig7}
  \end{figure*}

\subsection{Comparison with State-of-the-Art Methods}
We verify the effectiveness of our proposed method upon two sign language understanding tasks, i.e., continuous sign language recognition (CSLR) and sign language translation (SLT). We next introduce the results of our method upon both settings, respectively.

\begin{figure*}[t]
  \centering
  \includegraphics[width=0.84\linewidth]{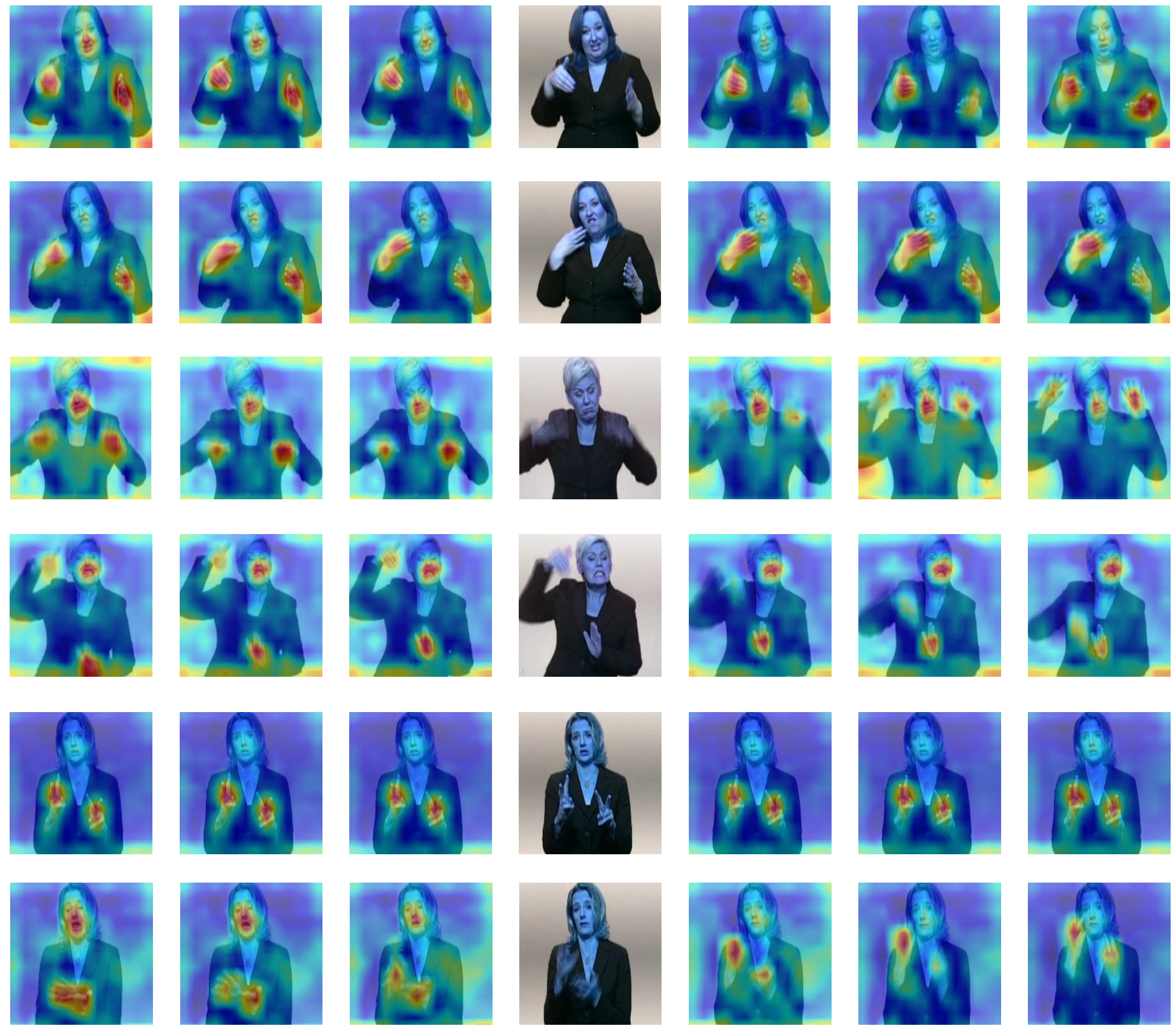}
  \caption{Visualizations of correlation maps for correlation module. Based on correlation operators, each frame could especially attend to informative regions in adjacent left/right frames like hands and face (dark red areas).}
  \label{fig6}
  \end{figure*}  

\begin{figure*}[!h]
  \centering
  \includegraphics[width=0.85\linewidth]{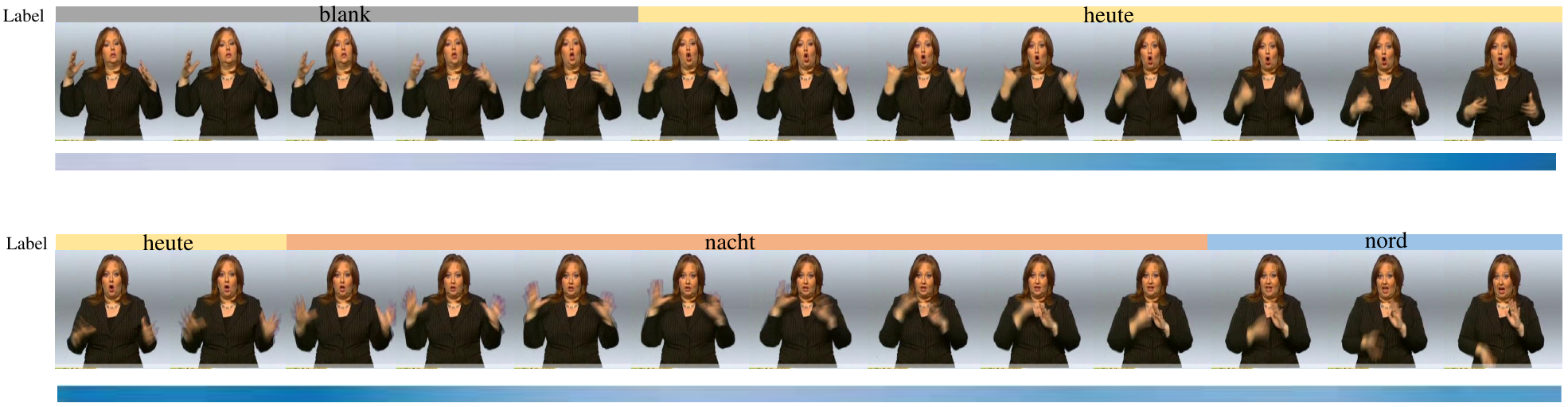}
  \caption{Visualizations of temporal attention maps for temporal attention module. It's observed that it tends to emphasize frames with rapid movements and suppress those frames with static contents.}
  \label{fig8}
  \end{figure*} 

\subsubsection{Continuous sign language recognition}
\textbf{PHOENIX2014} and \textbf{PHOENIX2014-T}. Tab.~\ref{tab8} shows a comprehensive comparison between our CorrNet+ and other state-of-the-art methods. The entries notated with $*$ indicate these methods utilize additional factors like face or hand features for better accuracy. We notice that CorrNet+ outperforms other state-of-the-art methods by a large margin upon both datasets, thanks to its special attention on body trajectories. Especially, CorrNet+ outperforms previous CSLR methods~\cite{koller2019weakly,zhou2020spatial,zuo2022c2slr,cui2019deep} equipped with hand and faces acquired by heavy pose-estimation networks or pre-extracted heatmaps (notated with *), without additional expensive supervision. 

\textbf{CSL-Daily}. CSL-Daily is a recently released large-scale dataset with the largest vocabulary size (2k) among commonly-used CSLR datasets, with a wide content covering family life, social contact and so on. Tab.~\ref{tab9} shows that our CorrNet+ achieves new state-of-the-art accuracy upon this challenging dataset with notable progress, which generalizes well upon real-world scenarios.

\textbf{CSL}. As shown in Tab.~\ref{tab10}, our CorrNet+ could achieve extremely superior accuracy (0.7\% WER) upon this well-examined dataset, outperforming existing CSLR methods.

\subsubsection{Sign language translation}
We compare our method with recent methods upon two widely-used SLT datasets, \textbf{Phoenix-2014T} and \textbf{CSL-Daily}, in Tab.~\ref{tab14}. These methods are roughly divided into two categories, \textit{Sign2Gloss2Text} which first transforms input videos into intermediate gloss representations and then performs translation, and \textit{Sign2Text} which directly conducts end-to-end translation from input videos. We observe that our method outperforms previous methods across both datasets, demonstrating its effectiveness in sign language comprehension. Especially, the powerful TwoStream-SLT~\cite{chen2022two} adopts both RGB videos and skeleton data as inputs to fuse beneficial information from both modalities, which requires more expensive supervision and heavy computations. In contrast, our method achieves better performance by only inputting RGB videos, demonstrating a better accuracy-computation trade-off.

\subsection{Visualizations}
\vspace{-2px}
\textbf{Visualizations for identification module.} Fig.~\ref{fig7} shows the heatmaps generated by our identification module. Our identification module pays special attention to the human body (light yellow areas), especially informative regions of hands and face (dark red areas) to capture human body trajectories. These results verify the effectiveness of our identification module in dynamically emphasizing critical areas in expressing sign language and suppressing other background regions to overlook noisy information.

\textbf{Visualizations for correlation module.} Fig.~\ref{fig6} illustrates the correlation maps generated by our correlation module, which shows the computed spatial-temporal correlations between the current frame and temporal neighboring frames. Three adjacent frames are shown to visualize the correlation maps. We observe that our correlation module pays major attention to informative regions in adjacent frames like hands or the face to enable precise tracking of body trajectories during sign expression. Especially, it learns to focus on the moving body parts that play a major role in expressing signs to enhance sign language comprehension. For example, in the 3rd and 4th row, the correlation module consistently pays major attention to the quickly moving right hands to capture sign information while overlooking the redundant information in the background.

\textbf{Visualizations for temporal attention module.} Fig.~\ref{fig8} visualizes the temporal attention maps generated by our temporal attention module over some selected frames. The darker color, the higher value.  We observe that our temporal attention module tends to allocate higher weights for frames with rapid movements (e.g., the latter several frames in the first line; the frontal frames in the second line). It learns to assign lower weights for static frames with few body movements. This observation is consistent with the habits of our human beings, as our humans always pay more attention to those moving objects in the visual field to capture key movements. These observations clearly reveal the effectiveness of our temporal attention module in emphasizing the critical segments in the whole sign video.
\section{Conclusion}
Recent methods on sign language understanding usually solely focus on each frame to extract their spatial features and overlook their cross-frame interactions, thus failing to capture the key human body movements. To handle this problem, this paper introduces an enhanced correlation network (CorrNet+) to capture human body trajectories, which comprises a correlation module, an identification module and a temporal attention module. The effectiveness of CorrNet+ is verified on two sign language understanding tasks including continuous sign language recognition (CSLR) and sign language translation (SLT) with new state-of-the-art performance compared to previous methods. Especially, by only inputting RGB videos on both tasks, CorrNet+ outperforms previous methods equipped with resource-intensive pose estimation networks or pre-extracted heatmaps with much fewer computations for hand and facial feature extraction. Compared to CorrNet~\cite{hu2023continuous}, CorrNet+ achieves a significant performance boost across multiple benchmarks with drastically reduced computational costs, demonstrating a better accuracy-computation trade-off. Plentiful visualizations further verify the effectiveness of CorrNet+ in intelligently emphasizing human body trajectories across adjacent frames in a self-motivated way. 

{
\bibliographystyle{IEEEtran}
\bibliography{ref}
}

\begin{IEEEbiography}[{\includegraphics[width=1in,height=1.25in,clip,keepaspectratio]{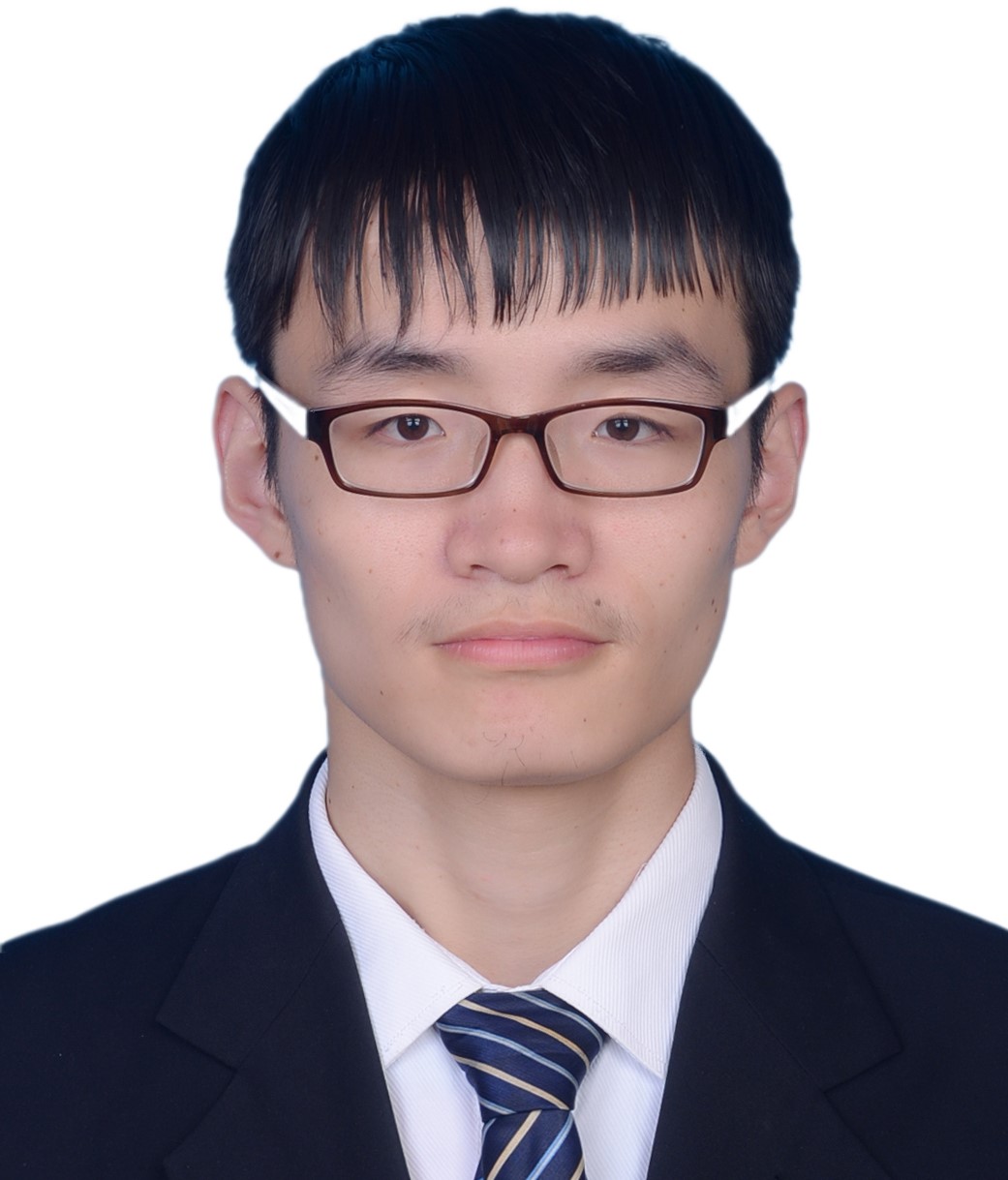}}]{Lianyu Hu} received the bachelor degree and master degree in computer science and technology from Dalian University of Technology in 2018 and 2021, respectively. He is currently a Ph.D. candidate with the College of Intelligence and Computing at Tianjin University, China. His research interests include video understanding, multimodal understanding and sign language understanding.
\end{IEEEbiography}

\begin{IEEEbiography}[{\includegraphics[width=1in,height=1.25in,clip,keepaspectratio]{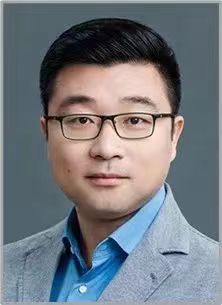}}] {Wei Feng} is a full Professor at the School of Computer Science and Technology, College of Computing and Intelligence, Tianjin University, China. He received the PhD degree in computer science from City University of Hong Kong in 2008. His major research interests are active robotic vision and visual intelligence, specifically including active camera relocalization and lighting recurrence, active 3D scene perception, video analysis, and generic pattern recognition.
\end{IEEEbiography}

\begin{IEEEbiography}[{\includegraphics[width=1in,height=1.25in,clip,keepaspectratio]{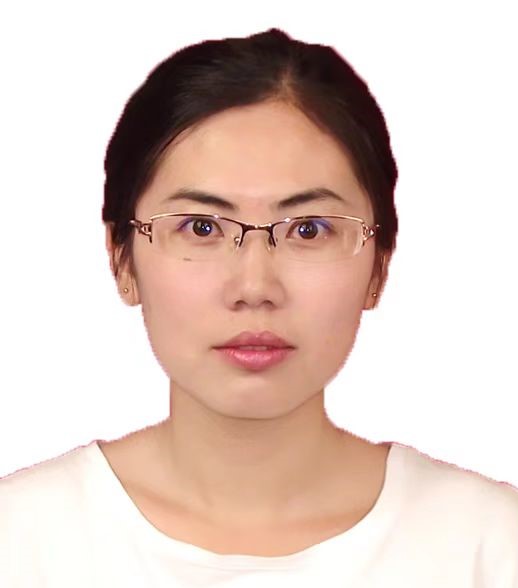}}]{Liqing Gao} received the BS and MS degree in Electronic \& Information Engineering, Inner Mongolia University, China, in 2015 and 2018. She is working toward the PhD degree in the College of Intelligence and Computing, Tianjin University, China. Her research interests include sign language recognition and gesture recognition.
\end{IEEEbiography}

\begin{IEEEbiography}[{\includegraphics[width=1in,height=1.25in,clip,keepaspectratio]{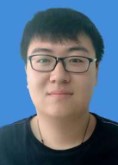}}]{Zekang Liu} received the BS and ME in Software Engineering from Hebei University of Economics and Business, China and Tianjin Normal University, China, in 2017 and 2019, respectively. He is studying for a Eng.D in the College of Intelligence Computing, Tianjin University, China. His research interests include vehicle detection and sign-language recognition.
\end{IEEEbiography}

\begin{IEEEbiography}[{\includegraphics[width=1in,height=1.25in,clip,keepaspectratio]{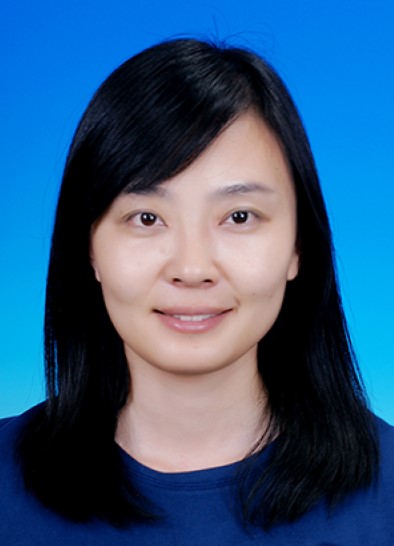}}]{Liang Wan} is a full Professor in the College of Intelligence Computing, and deputy director of Medical College, Tianjin University, P. R. China. She obtained a Ph.D. degree in computer science and engineering from The Chinese University of Hong Kong in 2007, and worked as a PostDoc Research Associate/Fellow at City University of Hong Kong from 2007 to 2011. Her current research interests focus on image processing and computer vision, including image segmentation, low-level image restoration, and medical image analysis.
\end{IEEEbiography}
\end{document}